\documentclass[]{style}

\usepackage[toc,page,header]{appendix}

\usepackage{xargs}

\usepackage{todonotes}

\usepackage{amsmath}
\usepackage{dsfont}

\usepackage{mathrsfs}
\usepackage{adjustbox}
\usepackage{multicol}
\usepackage{changepage}
\usepackage{enumitem}
\usepackage{amssymb}
\usepackage{float}
\usepackage{threeparttable}
\usepackage{algorithm}
\usepackage{algpseudocode}
\usepackage{wrapfig}
\usepackage{colortbl}       
\usepackage{array}
\usepackage{tabularx} 
\usepackage{makecell} 

\newcolumntype{g}{>{\columncolor{gray!10}}c} 

\definecolor{catgray}{gray}{0.9}
\definecolor{skyblue}{rgb}{0.53,0.81,0.92} 

\colorlet{skyblue!30}{skyblue!30!white} 

\definecolor{customblue}{RGB}{70,130,180}  

\newtcolorbox{evolbox}[2][]{%
  enhanced,
  colframe=customblue,
  colback=white,
  coltitle=white,
  rounded corners,
  boxrule=1pt,
  titlerule=0pt,
  toptitle=1mm,
  bottomtitle=1mm,
  fonttitle=\bfseries,
  width=#2\textwidth, 
  #1
}

\usepackage{minitoc}
\usepackage{url}

\usepackage{titletoc}
\usepackage{pifont}

\definecolor{RowBlue}{HTML}{E9F2FB}
\definecolor{RowRed}{HTML}{F9EAEA}
\definecolor{Top1}{HTML}{50DB4B} 
\definecolor{Top2}{HTML}{A5FFA2} 
\definecolor{Top3}{HTML}{D9FFD9} 
\definecolor{Sub1}{HTML}{EAB8B8}
\definecolor{Sub2}{HTML}{E4E4E4}

\renewcommand{\emph}[1]{\textit{#1}}

\definecolor{linkblue}{RGB}{76,104,191}
\definecolor{ourgray}{HTML}{F3F4F6}
\definecolor{deepred}{RGB}{180,0,0}
\definecolor{teal}{RGB}{0,128,128}

\title{Motion on Demand:\\Adaptive Optical Excitation for Event Cameras}

\author[1,2,*]{Yingtao~Li}
\author[1,*]{Chaoran~Feng}
\author[1,2,*]{Haoran~Jiang}
\author[1]{Zhenyu~Tang}
\author[1]{Wangbo~Yu}
\author[3]{Zhankuo~Xu}
\author[1]{Jianbin~Zhao}
\author[1,\dagger]{Li~Yuan}
\author[1,\dagger]{Yonghong~Tian}

\affiliation[1]{Shenzhen Graduate School, Peking University}
\affiliation[2]{School of Future Technology, Dalian University of Technology}
\affiliation[3]{School of Artificial Intelligence, Shanghai Jiao Tong University}

\abstract{

Event cameras respond only to brightness changes, so whether an edge is visible depends on image motion perpendicular to it---a motion that constantly shifts with camera and scene dynamics. Rotating a wedge prism in front of the lens introduces controlled image motion to reveal static or slow edges. However, a fixed rotation speed cannot adapt to changing motion conditions: when natural motion is already sufficient, additional optical excitation generates redundant events and introduces unnecessary alignment errors. Inspired by the division of labor between fast haltere feedforward and slower visual feedback in fly gaze control, we present EvGaze, an event camera system that adaptively regulates prism rotation speed. EvGaze employs a learned multimodal controller that integrates luminance, event, and inertial observations, with a fast inertial pathway that refines speed decisions between visual updates. To preserve geometric consistency during speed changes, we compensate for prism-induced image motion at individual event timestamps using measured encoder angles. On EvMotion, EvGaze recovers more edges than full-speed prism rotation with 39\% fewer events. On DSEC with simulated optical modulation, it reaches 0.73 feature-tracking success (compared with 0.55 on native events and 0.67 at 180 RPM) while generating 26\% fewer events than 360 RPM. Real-device experiments further demonstrate online closed-loop adaptation, reducing average event rates by 17\% relative to full-speed rotation.
}

\checkdata[
\raisebox{-0.2em}{\includegraphics[width=0.022\linewidth]{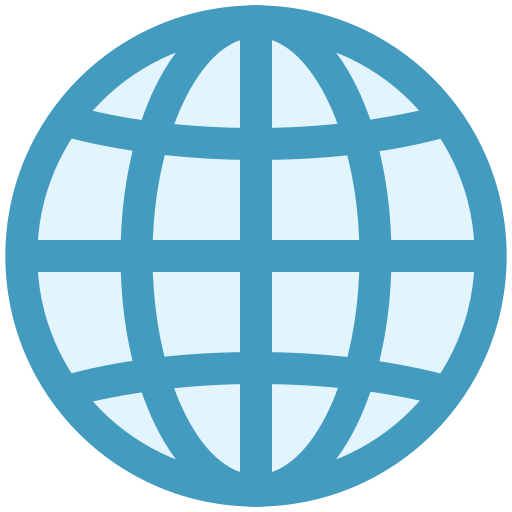}}~~Project Page]{\href{https://passwerob.github.io/EvGaze/}{\texttt{https://passwerob.github.io/EvGaze/}}}
\begin{document}
\maketitle

\begingroup
\renewcommand{\thefootnote}{\fnsymbol{footnote}}
\footnotetext[1]{These authors contributed equally to this work. Y. Li and H. Jiang are interns in the YuanGroup at Peking University.}
\footnotetext[2]{Corresponding author.}
\endgroup

\begin{figure}[ht]
    \centering
    \includegraphics[width=\linewidth,pagebox=cropbox]{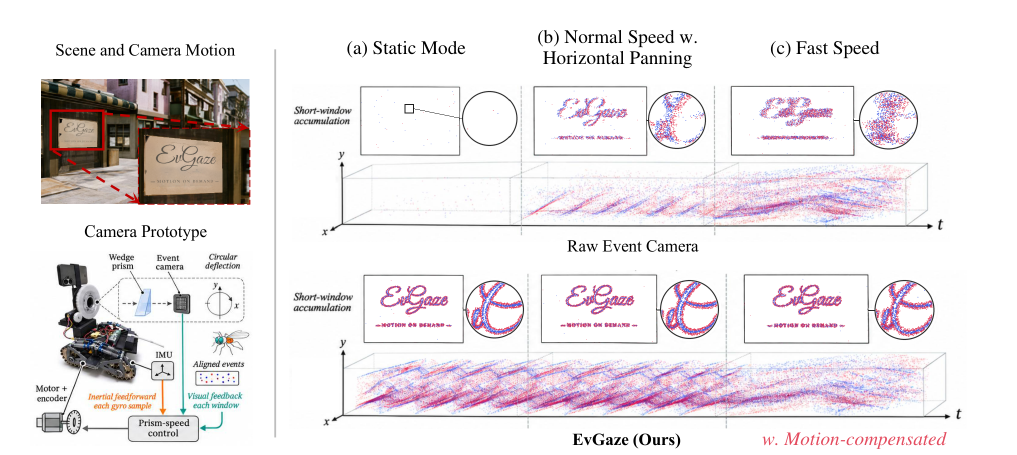}
    \caption{\textbf{Motion-adaptive event sensing with EvGaze.}
    A raw event camera sees little when static and misses edges parallel to its panning; only fast motion reveals the full scene. \textbf{EvGaze} adds prism motion only where natural motion falls short, guided by inertial feedforward and visual feedback, and recovers complete structure in all three cases.}
    \label{fig:teaser}
\end{figure}

\section{Introduction}
\label{sec:introduction}

Event cameras asynchronously record per-pixel log-intensity changes with microsecond resolution, high dynamic range, and low latency~\citep{lichtsteiner2008dvs,brandli2014davis,gallego2022event}, supporting high-speed feature tracking~\citep{gehrig2020eklt,messikommer2023data}, optical flow and event representation learning~\citep{zhu2018evflownet,gehrig2019est}, visual odometry and {SLAM}~\citep{zhu2017eventvio,mueggler2018continuous,rosinol2018ultimate,klenk2023devo}, motion segmentation and moving-object tracking~\citep{mitrokhin2018event,stoffregen2019event}, and 3D scene reconstruction, novel-view synthesis, and visual or embodied modeling~\citep{rudnev2023eventnerf,low2023robust,feng2025ae,yu2025evagaussians,yu2026ea3d,tang2026neuralgs,zhao2025tune,yu2025nofa++,tang2025cycle3d,zhou2026uniworld,yu2026worldcrafter,ma2026scaling,tang2026enhancing,zhao2026style,wu2026exfold}, including dynamic scenes~\citep{feng20254dgs} and motion-blurred captures~\citep{pan2019edi,feng2025d2gs,feng2026breaking,shi2026deblurnvs}. Because pixels fire only upon brightness changes, whether an edge is visible depends on its motion on the image plane---more specifically, on the motion component perpendicular to the edge~\citep{gallego2022event,he2024microsaccade}. In practice, this motion is driven by the platform and the scene, and constantly switches across states: a platform may hover, pan along a single axis (leaving edges parallel to the motion invisible), follow a target at matching speed (where the target barely moves on the image plane), or execute multi-directional maneuvers. When relative motion is weak or restricted to a single direction, some edges generate few or no events. Adaptive temporal windows can accumulate sparse responses for optical flow or feature tracking~\citep{liu2018adaptive,kim2026gostop}, but they cannot recover an edge that triggers no events: its brightness change stays below the contrast threshold however long the window is~\citep{gallego2022event}. Feature tracking, pose estimation, and object tracking all rely on continuously visible structure, and these intervals are precisely where they fail (\cref{fig:teaser}).

Mounting a rotating wedge prism in front of the lens introduces approximately circular image motion, allowing edges to trigger events even in a stationary scene. Encoder angles matched to event timestamps allow the induced shift to be compensated event by event~\citep{he2024microsaccade}. In AMI-EV, faster prism rotation generated more events and increased compensation errors, while the benefit of adding optical motion diminished as scene motion increased. 
This trade-off motivates adapting prism speed online. Existing methods regulate event generation through sensor-bias adjustment~\citep{delbruck2021feedback,nair2024enhancing} or filter events after acquisition~\citep{pellerito2026suppression}. 

A rotating prism acts on optical inputs, so it can generate events from edges that would otherwise remain unobserved. 
The challenge of online optical regulation lies in what signal to drive it with. Event feedback reflects what structure is observed, but reliable statistics take time to accumulate. 
A gyroscope immediately measures the strength and directional distribution of rotational ego-motion, yet knows nothing about object motion in the scene. One signal is fast but incomplete; the other is outcome-aware but slow.

In fly gaze stabilization, halteres sense body rotation and provide fast feedforward signals for compensatory head movements, while compound-eye motion signals provide slower visual feedback~\citep{schwyn2011interplay}. Visual and haltere responses are tuned to different rotation speeds~\citep{sherman2003comparison}, and both signals reach neck motor neurons~\citep{huston2009nonlinear}. Fruit flies also use retinal muscles to stabilize retinal images during visual motion and make small saccades while viewing stationary scenes~\citep{fenk2022muscles}. This suggests a simple principle: fast signals act first, and slow signals correct, without waiting for each other. Guided by this principle, we present \textbf{EvGaze}, which treats prism speed as an online control variable adjusted immediately by inertial feedforward and corrected by event-based visual feedback. We borrow the fly's division of labor rather than its motion-cancellation goal: instead of canceling motion, the prism adds missing motion and stops once natural motion is sufficient.

EvGaze consists of a variable-speed rotating prism in front of an event camera, an external time-aligned RGB camera outside the prism path, and a rigidly mounted IMU. A learned slow pathway integrates recent luminance, event, and inertial observations to produce baseline speed logits, while an inertial fast pathway refines them using updated IMU measurements and cached multimodal features. The resulting commands are executed under actuator constraints. Meanwhile, prism-induced image motion is compensated at individual event timestamps using encoder-measured angles, preserving geometric alignment during speed transitions. Event acquisition continues independently of control updates, enabling continuous downstream perception. Across simulation, real-world driving sequences, and physical hardware, we evaluate EvGaze through edge observability and feature tracking, alongside event-based 3D reconstruction, demonstrating competitive perception quality at lower event rates than continuous full-speed prism rotation.

Our main contributions are as follows:
\begin{itemize}
    \item We formulate optical excitation strength as an online decision problem: how much motion to add depends on what natural motion already provides.
    \item We present EvGaze, combining variable-speed prism excitation, a dual-rate multimodal controller, and per-event compensation under changing speeds.
    \item We validate EvGaze in simulation and on real hardware, demonstrating improved edge observability, event-based feature tracking, and 3D reconstruction at reduced event rates.
\end{itemize}

\section{Related Work}
\label{sec:related-work}

\subsection{Active and Bio-Inspired Motion Sensing}

Because event cameras respond only to brightness changes, active sensing methods modify the visual input before event generation to reveal static or slow structure. Aside from modulating optical transmittance~\citep{bao2024temporal}, most approaches introduce controlled motion: early systems used pan--tilt mechanisms for prescribed saccades~\citep{orchard2015converting,mishra2017saccade,yousefzadeh2018active} or mimicked fixational eye movements~\citep{rucci2007miniature,kuang2012temporal} to turn spatial contrast into temporal events~\citep{testa2020fixational,testa2023active}. To avoid moving the entire camera, AMI-EV~\citep{he2024microsaccade} rotates a wedge prism in front of the lens with per-event compensation, while VibES~\citep{polizzi2025vibes} vibrates the camera with an eccentric motor and removes the estimated displacement online. However, these systems apply a prescribed trajectory or fixed speed (both AMI-EV and VibES leave scene-adaptive regulation to future work), regardless of the natural motion already present. By contrast, biological vision regulates retinal motion online: in fruit flies (\emph{Drosophila}), retinal muscles beneath stationary compound-eye lenses generate visual responses during fixation and stabilize retinal images during motion~\citep{fenk2022muscles}, while gaze stabilization pairs rapid mechanosensory feedforward from the halteres with slower visual feedback~\citep{schwyn2011interplay}. Beyond optical mechanisms, bio-inspired spiking neural computation has also been explored to process asynchronous visual signals efficiently~\citep{shao2023eicil}. Whereas robotic frame cameras combine vestibulo-ocular and optokinetic reflexes to cancel ego-motion~\citep{habra2017gaze}, canceling ego-motion before an event camera would erase stationary scene structure, and constant strong excitation inflates event rates and alignment errors once natural motion is sufficient~\citep{gallego2022event,he2024microsaccade}. Motivated by this contrast, \textbf{EvGaze} regulates optical motion online before the sensor through fast inertial refinement and multimodal visual--event feedback.

\subsection{In-Sensor and Post-Acquisition Regulation}

Excessive event rates cause readout saturation, timestamp jitter, and dropped events~\citep{gehrig2022highres}. Inside the sensor, feedback controllers adjust contrast thresholds, refractory periods, and bandwidth from event rate and noise~\citep{delbruck2021feedback}, or use fast and slow loops to adapt the refractory period and remaining biases separately~\citep{nair2024enhancing}. Although Nair et al.~\citep{nair2024enhancing} also employ two timescales, both loops are driven solely by scalar event rates and only alter sensitivity to existing brightness changes; EvGaze instead fuses inertial, luminance, and event cues to regulate the optical motion entering the sensor. After acquisition, software methods suppress predicted redundant events~\citep{pellerito2026suppression} or derotate event streams using IMU measurements~\citep{delbruck2014integration,rodriguez2024benefits}, echoing broader efforts in selective visual filtering and concept suppression~\citep{li2026fade}. For event accumulation, contrast maximization warps events to align edges~\citep{gallego2018contrast} and segment moving objects~\citep{stoffregen2019event}; because weak relative motion yields insufficient displacement in short windows~\citep{stoffregen2019event}, adaptive methods adjust window lengths~\citep{liu2018adaptive} or learn when to stop accumulating (GoStop~\citep{kim2026gostop}). Complementing sensor-level event simulators~\citep{rebecq2018esim,hu2021v2e,lin2022dvsvoltmeter}, recent generative approaches also synthesize event streams directly from 3D Gaussian Splatting~\citep{li2026gs2e} or leverage large-scale visual and video generation backbones~\citep{lin2024open,pang2024next} together with structured semantic evaluation~\citep{niu2025wise} and unified multimodal understanding~\citep{niu2025does}. While in-sensor and post-acquisition methods suppress redundancy or reorganize emitted events, they cannot recover edges whose brightness changes remain below threshold. EvGaze supplies missing optical motion before event generation, operating at a stage orthogonal and complementary to these methods.

\section{Method}
\label{sec:method}

\begin{figure}[t]
    \centering
    \includegraphics[width=\linewidth]{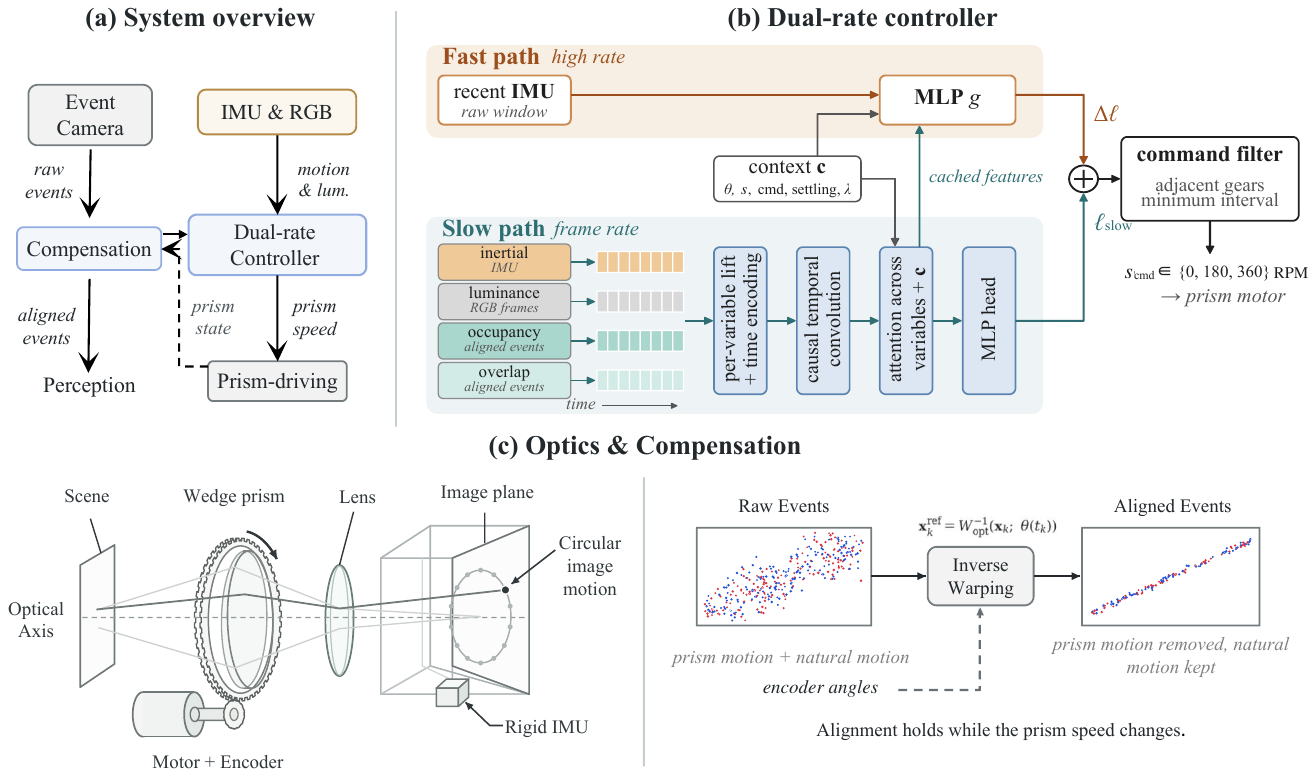}
    \caption{\textbf{Overview of EvGaze.} (a) System overview integrating event, inertial, and external RGB streams. (b) Dual-rate controller with a multimodal slow pathway and an inertial fast pathway. (c) Prism-based optical excitation and per-event encoder-angle compensation.}
    \label{fig:pipeline}
\end{figure}

\textbf{EvGaze} formulates active event sensing as a closed-loop rate--distortion control problem, optimizing wedge-prism rotation to balance structural observability and event bandwidth. The system operates across three timescales: continuous optical modulation with timestamp-based event alignment, frame-rate multimodal temporal encoding, and high-rate IMU-driven control under actuator constraints. We first introduce optical modulation and actuator dynamics (\cref{sec:optical_physics}), then present the dual-rate controller (\cref{sec:controller}), followed by the Target Event Observability (TEO) criterion and latency-aware training (\cref{sec:dp_training}).

\subsection{Optical Modulation and Actuator Physics}
\label{sec:optical_physics}

\paragraph{Optical Modulation Before Event Generation.}
An event camera triggers at pixel $\mathbf{x}$ only when the temporal log-intensity change reaches a contrast threshold $C_{\mathrm{th}}$, i.e., $|\Delta \log I(\mathbf{x}, t)| \ge C_{\mathrm{th}}$. Under linearized brightness constancy, the instantaneous intensity rate follows
\begin{equation}
\frac{\partial \log I(\mathbf{x}, t)}{\partial t} \approx -\nabla \log I(\mathbf{x}, t) \cdot \mathbf{v}_{\text{img}}(\mathbf{x}, t),
\label{eq:event_physics}
\end{equation}
\noindent
\begin{minipage}[c]{0.50\linewidth}
where $\mathbf{v}_{\text{img}}$ is the apparent image velocity. When a scene is stationary, under weak image motion, or moving parallel to an edge ($\nabla \log I \perp \mathbf{v}_{\text{img}}$), edge-triggered event activity becomes weak or vanishes. Post-hoc filtering or adaptive windowing cannot recover structures whose sub-threshold changes are never recorded by the sensor. As illustrated in \cref{fig:optical_physics}, placing a rotating wedge prism in front of the lens alters image formation before photoreception by adding a known optical velocity $\mathbf{v}_{\text{opt}}(\mathbf{x}, t)$, yielding $\mathbf{v}_{\text{eff}} = \mathbf{v}_{\text{img}} + \mathbf{v}_{\text{opt}}$.
\end{minipage}\hfill
\begin{minipage}[c]{0.47\linewidth}
    \centering
    \includegraphics[width=\linewidth]{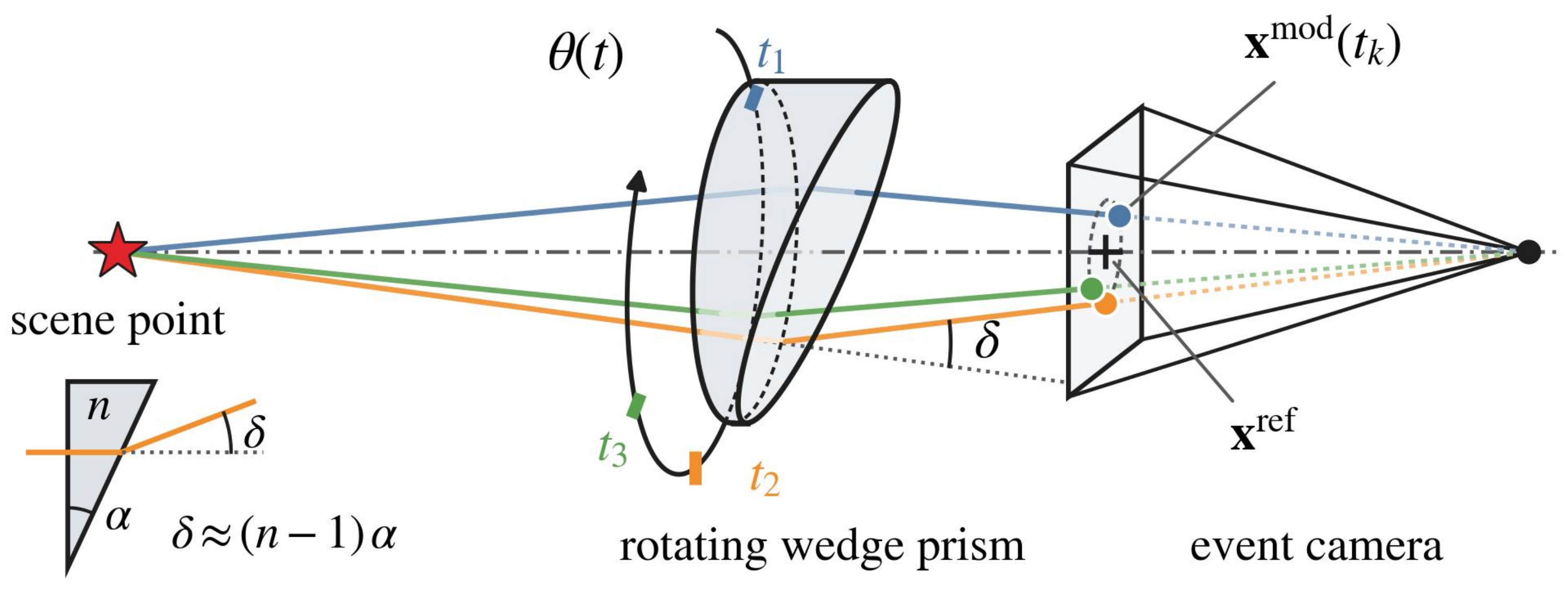}
    \captionof{figure}{\textbf{Pre-event excitation and timestamp alignment.}}
    \label{fig:optical_physics}
\end{minipage}

\noindent Unlike single-axis mechanical shaking, which still leaves parallel edges unobserved, a rotating prism sweeps every scene point along a two-dimensional circular path. Across each rotation cycle, the optical velocity spans all orientations, producing nonzero edge-normal motion over part of the cycle and thereby activating otherwise unobserved spatial gradients.

\paragraph{Optical Mapping and Per-Event Compensation.}
Let $\mathbf{K}$ be the camera intrinsic matrix with focal lengths $(f_x, f_y)$ and $\bar{\mathbf{x}} = [u, v, 1]^\top$ be the homogeneous pixel coordinate. For a wedge prism with apex angle $\alpha$ and refractive index $n$, the paraxial deflection $\delta \approx (n-1)\alpha$ at rotation phase $\theta(t)$ yields $\mathbf{q}(\theta) = \delta[\cos\theta, \sin\theta, 0]^\top$ and induces the closed-form image mapping:
\begin{equation}
W_{\text{opt}}(\mathbf{x}; \theta) = \mathbf{x} + \rho(\mathbf{x})\,\delta \begin{bmatrix} f_x \cos\theta \\ f_y \sin\theta \end{bmatrix}, \quad \rho(\mathbf{x}) = \|\mathbf{K}^{-1}\bar{\mathbf{x}}\|_2.
\label{eq:forward_warp}
\end{equation}
The sensor observes the modulated intensity field $S(\mathbf{x}, t) = I(W_{\text{opt}}^{-1}(\mathbf{x}; \theta(t)), t)$, from which events are generated (\cref{fig:optical_physics}). Under this analytical model, $W_{\text{opt}}$ admits a closed-form quadratic inverse $W_{\text{opt}}^{-1}$. Each recorded event $e_k=(\mathbf{x}_k^{\text{mod}},t_k,p_k)$ is mapped back to the reference image plane at its timestamp $t_k$:
\begin{equation}
\mathbf{x}_k^{\text{ref}} = W_{\text{opt}}^{-1}\big(\mathbf{x}_k^{\text{mod}};\, \theta(t_k)\big).
\label{eq:per_event_comp}
\end{equation}
In simulation, this analytical mapping is used for both pre-event image formation and inverse compensation. On our physical prototype, we calibrate the prism-induced displacement as a circular translation $\mathbf{d}_{\mathrm{hw}}(\theta_m)
=R[\cos(\phi_{\mathrm{hw}}-\theta_m),
\sin(\phi_{\mathrm{hw}}-\theta_m)]^\top$,
with radius $R$ and phase offset $\phi_{\mathrm{hw}}$ fitted by contrast maximization~\citep{gallego2018contrast}. Hardware events are compensated using the encoder phase at their individual timestamps, preserving alignment during speed transitions.

\paragraph{Actuator Dynamics and Switching Constraints.}
Real rotary drives exhibit finite inertia and communication latency rather than instantaneous step changes. Calibrating against motor encoder logs, we model the actual prism speed $s(t)$ via a first-order system with pure dead time:
\begin{equation}
\tau \dot{s}(t) = s_{\text{cmd}}(t - t_d) - s(t), \quad \theta(t) = \theta_0 + \frac{2\pi}{60}\int_0^t s(\xi)\,d\xi,
\label{eq:motor_dynamics}
\end{equation}
where $t_d$ is the actuation delay and $\tau$ is the motor time constant. To prevent mechanical chatter, commands $s_{\text{cmd}}(t) \in \mathcal{A}_p=\{0,180,360\}\,\mathrm{RPM}$ are restricted to adjacent gear transitions ($s_{\text{low}} \leftrightarrow s_{\text{mid}} \leftrightarrow s_{\text{high}}$) with a minimum dwell interval $\Delta t_{\min}$ between switches. These actuator constraints motivate our dual-rate controller (\cref{sec:controller}) and actuator-aware supervision (\cref{sec:dp_training}). The closed-form inverse derivation and measured hardware actuator responses are provided in \cref{supp:optics,supp:actuator}.

\subsection{Dual-Rate Closed-Loop Controller}
\label{sec:controller}

\paragraph{Causal Multimodal Sensing State.}
Event activity alone cannot reliably distinguish camera-induced motion from independently moving objects. When activity arises from a moving foreground object while the platform remains stationary, static background structures may still require optical excitation. Conversely, strong camera rotation may already excite much of the scene, reducing the need for additional prism motion. We therefore construct a causal sensing history on a uniform inertial time grid using gyroscope and accelerometer measurements, prism-independent log-luminance variation from the external RGB view, spatial occupancy statistics of prism-compensated events, and inter-window event-support overlap. These complementary cues characterize ego-motion dynamics, event activity, and temporal consistency without assuming that event activity alone identifies its motion source.

\paragraph{Factorized Temporal Slow Pathway.}
At $60\,\mathrm{Hz}$, the slow pathway aggregates multimodal sensing history (\cref{fig:pipeline}). Each input stream is projected into a shared latent space and processed independently along time by causal dilated convolutions with residual connections and GroupNorm, before cross-modal fusion. The latest modality tokens are combined with a context token $\mathbf{c}(t)$ encoding prism phase $(\sin\theta,\cos\theta)$, motor speed $s$, current gear command, transition status, and the externally specified bandwidth price $\lambda$. Multi-head self-attention followed by LayerNorm fuses the tokens without imposing an artificial spatial ordering. The four attended modality tokens form a cached representation $\mathbf{z}\in\mathbb{R}^{4d}$; an MLP head takes $[\mathbf{z},\mathbf{c}(t)]$ to produce baseline speed logits $\operatorname{head}_{\mathrm{slow}}(\mathbf{z},\mathbf{c}(t))$.

\paragraph{Inertial Fast Pathway.}
Platform motion may change between visual updates, introducing decision latency in addition to the actuator dead time $t_d$. A lightweight residual module $g(\cdot)$ therefore operates at the inertial update rate ($200\,\mathrm{Hz}$ in simulation and $100\,\mathrm{Hz}$ on our physical prototype) and refines the speed logits using the cached representation $\mathbf{z}$, recent inertial measurements $\mathbf{u}_{\mathrm{imu}}(t)$, the representation staleness $\mathrm{age}$, and the current actuator context $\mathbf{c}(t)$:
\begin{equation}
\boldsymbol{\ell}(t)
=\operatorname{head}_{\mathrm{slow}}\big(\mathbf{z},\mathbf{c}(t)\big)
+g\big(\mathbf{z},\mathbf{u}_{\mathrm{imu}}(t),\mathbf{c}(t),\mathrm{age}\big).
\label{eq:fast_pathway}
\end{equation}
Caching $\mathbf{z}$ rather than the slow logits allows the decision head and residual pathway to use updated motor states and prism phase at each inertial step, without rerunning the temporal backbone. This enables faster responses to measured ego-motion changes under actuator constraints.

\paragraph{Actuation Execution.}
An execution layer converts $\mathbf{p}=\operatorname{softmax}(\boldsymbol{\ell}(t))$ into admissible motor commands under \cref{eq:motor_dynamics}. A switch is accepted only if the candidate probability exceeds that of the current commanded gear by a hysteresis margin $\delta_p$; direct transitions between the extreme gears pass through the intermediate gear, and successive commands obey the minimum dwell interval $\Delta t_{\min}$. Dual-rate timing and command-execution details are discussed in \cref{supp:actuator,supp:controller}.

\subsection{Target Observability and Policy Learning}
\label{sec:dp_training}

\paragraph{Target Event Observability.}
Raw event activity does not establish whether useful target structure has been observed. We use Target Event Observability (TEO) to measure target-edge coverage, purity, and temporal persistence. Let $\mathcal{T}$ denote a target region, $G\subseteq\mathcal{T}$ its RGB-derived reference edges, and $M$ the spatial support of prism-compensated events. We define
\begin{equation}
\begin{gathered}
C=\frac{|G\cap M|}{|G|},\quad
P=\frac{|G\cap M|}{|M\cap\mathcal{T}|},\\[4pt]
\mathrm{TEO}=\frac{2CP}{C+P}\,\eta,
\end{gathered}
\label{eq:teo}
\end{equation}
where $\eta$ is the fraction of temporal subintervals containing target-edge observations. TEO is set to zero whenever $C+P=0$ or a required support is empty; windows without reference target edges are excluded from TEO evaluation. Coverage measures the observed fraction of target edges, while purity measures the fraction of in-target event support aligned with those edges. TEO is intentionally target-specific: background events do not lower its value. We therefore treat global structural fidelity and total event bandwidth as separate components of the training objective.

\paragraph{Quality--Bandwidth Objective.}
We balance structural observability and event bandwidth using matched fixed-speed rollouts. Observation quality is defined as $Q=(\mathrm{F1}+\mathrm{TEO})/2$, using F1 alone when target-level observability is unavailable. For each window $w$ and prism gear $g$, we define
\begin{equation}
J_{\lambda}(w,g)
=Q(w,g)-\lambda\frac{r(w,g)}{r_{\mathrm{ref}}},
\label{eq:rate_distortion}
\end{equation}
where $r(w,g)$ denotes the full-sensor event rate and $r_{\mathrm{ref}}$ is a fixed normalization scale. We use $\lambda=0.035$ for all main experiments.

\paragraph{Actuator-Aware Dynamic Programming Teacher.}
To account for command latency and motor inertia, we construct an offline dynamic-programming (DP) teacher incorporating the actuator dynamics in \cref{eq:motor_dynamics}, adjacent-gear constraints, and minimum command intervals. Motor responses are integrated over each window, with utility held at the previous gear during dead time and interpolated by instantaneous speed during ramping. The DP teacher computes action regrets $R_{\lambda}(a)$ relative to the optimal achievable future utility under the quality--bandwidth objective. Supervision covers both stable-gear states and intermediate switching trajectories, allowing the student to learn the effects of delayed actuation without explicitly simulating events during speed transitions.

\paragraph{Regret-Guided Policy Optimization.}
Let $\mathbf{o}_t$ denote the causal controller observations. We optimize the policy $\pi_{\psi}(a\mid\mathbf{o}_t,\lambda)$ using clipped expected regret, selective cross-entropy, and a gear-reversal penalty:
\begin{equation}
\begin{aligned}
\mathcal{L}
={}&\mathbb{E}_{a\sim\pi_{\psi}}
\left[\frac{\min(R_{\lambda}(a),\kappa)}{\kappa}\right] \\
&+\beta_{\mathrm{CE}}\mathcal{L}_{\mathrm{CE}}
+\beta_{\mathrm{cyc}}\mathcal{L}_{\mathrm{cycle}}.
\end{aligned}
\label{eq:policy_loss}
\end{equation}
Here $\kappa$ is the regret-clipping threshold. Cross-entropy favors the minimum-regret action when its advantage over the runner-up is large enough, while the cycle penalty discourages consecutive gear reversals. Training proceeds in two stages. We first train the slow pathway using teacher regrets. We then freeze the slow pathway and optimize the inertial residual module $g(\cdot)$ of the fast pathway using the same regret-based supervision at the IMU update rate. Additional supervision and control details are provided in \cref{supp:controller}.

\section{Experiments}
\label{sec:experiments}

We evaluate \textbf{EvGaze} across controlled motion, recorded driving, reconstructed real scenes, and physical acquisition. Our experiments ask whether optical excitation recovers otherwise weakly observed edges, whether adaptive excitation offers a favorable quality--event-rate operating point, and whether the resulting events support downstream tracking and reconstruction. We distinguish simulated counterfactual prism interventions from events acquired by the physical prototype throughout.

\subsection{Experimental Setup}
\label{sec:experimental_setup}

\paragraph{Datasets and Simulation.}
We evaluate EvGaze on three datasets. \textbf{EvMotion} is our Blender-based benchmark with controlled camera--object motion, covering four regimes: \emph{camera-only}, \emph{object-only}, \emph{co-motion}, and \emph{drive-stop-drive}. It contains multiple scene instances for training, validation, and testing, rendered at $960\!\times\!540$ and $60\,\mathrm{fps}$. Target masks and pose-derived $200\,\mathrm{Hz}$ IMU measurements support observability evaluation and policy control. On \textbf{DSEC}~\citep{gehrig2021dsec}, we use three driving sequences with recorded camera motion and IMU measurements. Prism-modulated events are simulated from temporally interpolated and aligned RGB sequences, while native events provide a real-sensor reference. We adopt leave-one-sequence-out fine-tuning. On \textbf{DL3DV}~\citep{ling2024dl3dv}, we synthesize events from RIFE-interpolated~\citep{huang2022rife} RGB sequences for event-only reconstruction with E2VID~\citep{Rebecq19cvpr} and 3DGS~\citep{kerbl20233dgs} as well as event-assisted motion-deblurring reconstruction~\citep{yu2025evagaussians} (\cref{supp:evagaussians_method}).

\paragraph{Physical Prototype.}
Our prototype integrates a motor-driven wedge prism in front of a $640\!\times\!480$ event camera, a $100\,\mathrm{Hz}$ IMU rigidly attached to the camera body (driving the inertial fast pathway online between visual updates), a prism-angle encoder, and an external time-aligned RGB camera mounted outside the prism optical path (\cref{supp:hardware}). We collect nine $15\,\mathrm{s}$ recordings across three motion conditions at fixed prism speeds of $0$, $180$, and $360\,\mathrm{RPM}$ to adapt the controller to the sensor domain, followed by online closed-loop deployment on the prototype. Acquired events are compensated using calibrated prism motion at their individual timestamps.

\paragraph{Baselines, Metrics, and Implementation.}
We compare EvGaze against passive simulators DVS-Voltmeter~\citep{lin2022dvsvoltmeter} and v2e~\citep{hu2021v2e}, alongside fixed optical excitation at $0$, $180$, and $360\,\mathrm{RPM}$. Here Fixed-180 and Fixed-360 instantiate constant-speed wedge-prism scanning---the fixed-speed paradigm of AMI-EV~\citep{he2024microsaccade}---within the exact same event-generation and per-event compensation pipeline for matched comparison. We report edge F1, TEO (\cref{eq:teo}), and event rate (Mev/s), together with feature tracking and 3D reconstruction quality. All main EvGaze experiments use $\lambda=0.035$. Event simulation and compensation are CUDA-accelerated.

\subsection{Optical Excitation and Control}
\label{sec:motion_observability}


\begin{table}[t]
\centering

\begin{minipage}[t]{0.49\textwidth}
\vspace{0pt}

\begin{minipage}[t][14mm][t]{\linewidth}
\caption{\textbf{Quantitative comparison on EvMotion.}
Average Edge F1, TEO, and event rates across four
motion regimes.}
\label{tab:main_combined}
\end{minipage}

\centering
\footnotesize
\setlength{\tabcolsep}{1.2pt}
\renewcommand{\arraystretch}{1.12}

\resizebox{\linewidth}{!}{%
\begin{tabular}{lcc ccccc}
\toprule
\multirow{2}{*}{Method}
& Edge F1$\uparrow$ & TEO$\uparrow$
& \multicolumn{5}{c}{Event rate (Mev/s)$\downarrow$} \\
\cmidrule(lr){2-2}
\cmidrule(lr){3-3}
\cmidrule(lr){4-8}
& Avg. & Avg. & Cam. & Obj. & Co. & DSD & Avg. \\
\midrule
DVS-Voltmeter~\cite{lin2022dvsvoltmeter}
& 0.275 & 0.281 & 10.8 & 0.4 & 9.8 & 7.7 & 7.2 \\
v2e~\cite{hu2021v2e}
& 0.311 & 0.293 & 48.9 & 1.6 & 44.6 & 32.8 & 32.0 \\
\midrule
Fixed-0
& 0.346 & 0.368 & 51.6 & 5.6 & 47.8 & 37.3 & 35.6 \\
Fixed-180
& \underline{0.548} & \underline{0.465}
& 69.4 & 41.6 & 70.1 & 66.9 & 62.0 \\
Fixed-360
& 0.529 & 0.441
& 101.7 & 78.9 & 106.1 & 105.8 & 98.1 \\
\midrule
\rowcolor{ourgray}
EvGaze (Ours)
& \textbf{0.553} & \textbf{0.466}
& 69.4 & 32.3 & 70.1 & 66.9 & 59.7 \\
\bottomrule
\end{tabular}%
}
\end{minipage}
\hfill
\begin{minipage}[t]{0.49\textwidth}
\vspace{0pt}

\begin{minipage}[t][14mm][t]{\linewidth}
\caption{\textbf{Target-level observability on EvMotion.}
TEO across four motion regimes. Best results in bold,
second-best underlined.}
\label{tab:teo_main}
\end{minipage}

\centering
\footnotesize
\setlength{\tabcolsep}{3pt}
\renewcommand{\arraystretch}{1.12}

\resizebox{\linewidth}{!}{%
\begin{tabular}{l cccccc}
\toprule
\multirow{2}{*}{Method}
& \multicolumn{5}{c}{TEO$\uparrow$}
& \multirow{2}{*}{Mev/s$\downarrow$} \\
\cmidrule(lr){2-6}
& Cam. & Obj. & Co. & DSD & Avg. & \\
\midrule
DVS-Voltmeter~\cite{lin2022dvsvoltmeter}
& \textbf{0.340} & 0.214 & 0.317 & 0.252 & 0.281 & 7.2 \\
v2e~\cite{hu2021v2e}
& 0.266 & 0.299 & 0.344 & 0.265 & 0.293 & 32.0 \\
\midrule
Fixed-0
& 0.301 & 0.470 & 0.375 & 0.326 & 0.368 & 35.6 \\
Fixed-180
& \underline{0.317} & \underline{0.546}
& \textbf{0.521} & \textbf{0.477}
& \underline{0.465} & 62.0 \\
Fixed-360
& 0.311 & 0.501 & \underline{0.504}
& \underline{0.448} & 0.441 & 98.1 \\
\midrule
\rowcolor{ourgray}
EvGaze (Ours)
& \underline{0.317} & \textbf{0.549}
& \textbf{0.521} & \textbf{0.477}
& \textbf{0.466} & 59.7 \\
\bottomrule
\end{tabular}%
}
\end{minipage}

\end{table}

\paragraph{Optical Excitation and Event Efficiency.}
We first evaluate structural observability and event cost on EvMotion (\cref{tab:main_combined}). Active optical excitation substantially improves edge recovery compared with passive event generation, but stronger excitation does not always provide better observability. Increasing prism speed from 180 to 360 RPM increases event activity while reducing both Edge F1 and TEO. By adjusting excitation to the observed motion, EvGaze achieves the highest average observability while producing fewer events than Fixed-180. These results favor adaptive control over constant high-speed excitation.

\paragraph{Motion-Regime Analysis.}
To explain these aggregate results, we examine target-level observability across motion regimes (\cref{tab:teo_main}). Unlike whole-image Edge F1, TEO evaluates target coverage, purity, and temporal persistence using target masks and RGB-derived references. Active excitation is particularly useful under co-motion and drive-stop-drive, where natural image motion is insufficient or intermittent. Under object-only motion, the moving target is already observable without added excitation, whereas the background benefits from optical motion. EvGaze thus maintains target observability comparable to Fixed-180 with fewer events, especially under object-only motion.

\begin{figure}[t]
    \centering
    \includegraphics[width=\linewidth]{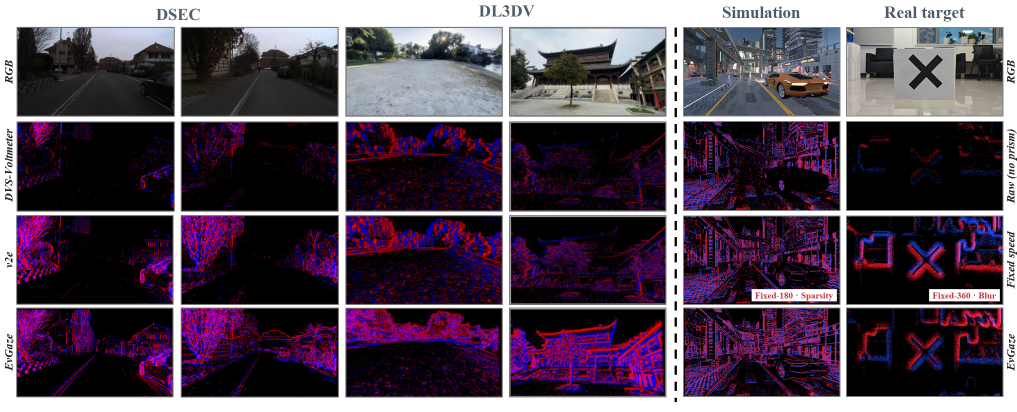}
    \caption{
    \textbf{Qualitative results across DSEC, DL3DV, simulation, and real hardware.}
    The left four columns (\emph{DSEC}, \emph{DL3DV}) compare EvGaze with passive event simulators (DVS-Voltmeter and v2e), while the right two columns (\emph{Simulation}, \emph{Real target}) compare against unmodulated acquisition (Raw) and constant-speed scanning (Fixed-180 in simulation, Fixed-360 on hardware).
    Fixed-180 leaves sparse edges under slow motion, whereas Fixed-360 introduces residual blur around the physical target; EvGaze recovers missing edges without sacrificing sharpness.
    Red and blue denote event polarities over matched accumulation windows.
    }
    \label{fig:qualitative_results}
\end{figure}

\paragraph{Qualitative Observations.}
Qualitative comparisons on DSEC and DL3DV (\cref{fig:qualitative_results}) further illustrate the effect of optical excitation. Compared with DVS-Voltmeter and v2e, EvGaze reveals additional structural responses in regions with limited natural motion. These observations complement the quantitative results by showing where optical excitation increases event visibility.

\subsection{Downstream Perception}
\label{sec:downstream_perception}
\label{sec:tracking_evaluation}

We next evaluate whether improved event observability translates into useful downstream perception, considering feature tracking and 3D reconstruction on synthetic and real-world sequences.

\begin{figure}[t]
    \centering
    \includegraphics[width=\linewidth]{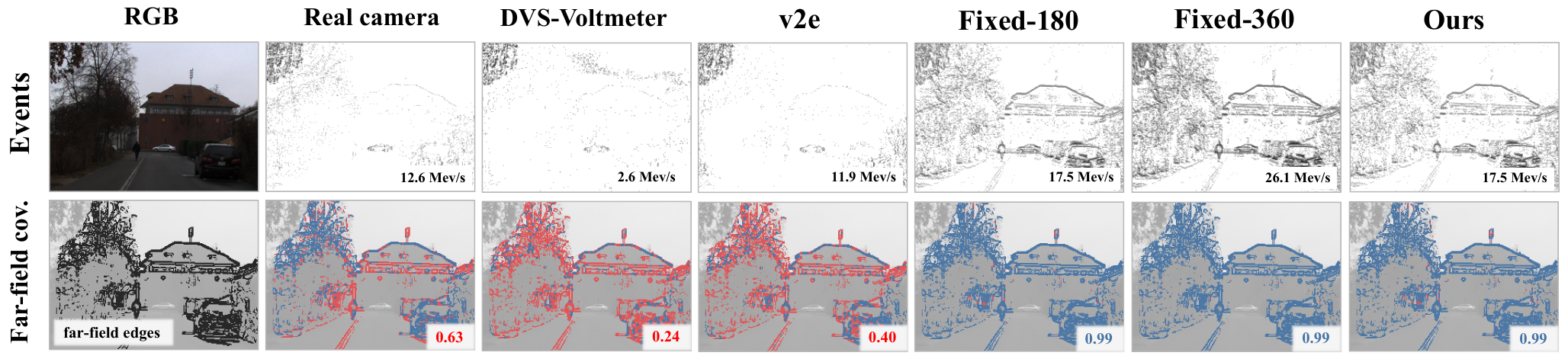}
    \caption{\textbf{Event-based perception on DSEC.} Representative $16.7\,\mathrm{ms}$ event visualizations (top) and far-field edge coverage (bottom; blue: covered reference edges, red: missed edges). Overlaid numbers denote single-window event rate and coverage on this low-motion interval where EvGaze selects $180\,\mathrm{RPM}$; sequence-level averages are reported in \cref{tab:tracking_all}.}
    \label{fig:dsec_tracking}
\end{figure}

\begin{table}[t]
\centering
\caption{\textbf{Feature tracking on EvMotion and DSEC.}
EvGaze is fine-tuned on DSEC using leave-one-sequence-out
evaluation; ZS denotes zero-shot transfer.
Rates are in Mev/s. Best results in bold,
second-best underlined.}
\label{tab:tracking_all}

\small
\setlength{\tabcolsep}{2.5pt}
\renewcommand{\arraystretch}{1.08}
\resizebox{\linewidth}{!}{
\begin{tabularx}{0.96\textwidth}{
l
*{5}{>{\centering\arraybackslash}X}
>{\centering\arraybackslash}p{58pt}
*{3}{>{\centering\arraybackslash}X}
}
\toprule
\multirow{2}{*}{Source}
& \multicolumn{4}{c}{EvMotion}
& \multicolumn{5}{c}{DSEC} \\
\cmidrule(lr){2-5}
\cmidrule(lr){6-10}
& Rate
& S@1$\uparrow$
& S@3$\uparrow$
& Med.$\downarrow$
& Rate
& Far-field cov.$\uparrow$
& S@1$\uparrow$
& S@3$\uparrow$
& Med.$\downarrow$ \\
\midrule

Real event camera
& -- & -- & -- & --
& 12.7 & 0.621 & 0.554 & 0.929 & 1.16 \\

DVS-Voltmeter
& 7.2 & 0.267 & 0.773 & 2.93
& 1.7 & 0.213 & 0.299 & 0.791 & 5.84 \\

v2e
& 32.0 & 0.327 & 0.803 & 3.27
& 6.5 & 0.424 & 0.315 & 0.718 & 13.60 \\

\midrule

Fixed-0
& 35.6 & 0.343 & 0.814 & 4.14
& 6.6 & 0.450 & 0.314 & 0.718 & 14.57 \\

Fixed-180
& 62.0 & 0.561 & 0.917 & 1.18
& 12.0 & 0.944 & 0.665 & 0.951 & 1.18 \\

Fixed-360
& 98.1
& \textbf{0.649}
& \textbf{0.936}
& \textbf{0.94}
& 20.8
& \textbf{0.979}
& \textbf{0.774}
& \textbf{0.987}
& \textbf{0.71} \\

\midrule

EvGaze (ZS)
& -- & -- & -- & --
& 10.0 & 0.790 & 0.545 & 0.887 & 1.23 \\

\rowcolor{ourgray}
EvGaze (Ours)
& 59.7
& \underline{0.578}
& \underline{0.927}
& \underline{1.13}
& 15.4
& \underline{0.970}
& \underline{0.733}
& \underline{0.969}
& \underline{1.00} \\

\bottomrule
\end{tabularx}
}
\end{table}

\paragraph{Event-Based Feature Tracking.}
We evaluate feature tracking on EvMotion and three real-world DSEC driving sequences (\cref{tab:tracking_all}). On EvMotion, EvGaze achieves higher tracking accuracy than Fixed-180 at a comparable event rate. On DSEC, leave-one-sequence-out fine-tuning improves S@1 from 0.545 to 0.733 and far-field edge coverage from 0.790 to 0.970. Compared with Fixed-360, EvGaze maintains competitive tracking accuracy while generating 26\% fewer events. As visualized in \cref{fig:dsec_tracking}, optical excitation substantially improves far-field edge observability. All simulated DSEC sources share the same registered RGB sequences and camera trajectories, while native events are recorded by the physical event camera.





\begin{table}[t]
\centering
\caption{\textbf{Event-only reconstruction with E2VID + 3DGS.}
Results on DL3DV and DSEC using event-reconstructed intensity images. Best in bold, second best underlined.}
\label{tab:e2vid_3dgs}
\footnotesize
\setlength{\tabcolsep}{3.5pt}
\renewcommand{\arraystretch}{1.12}
\begin{minipage}[t]{0.44\textwidth}
\centering
\begin{tabularx}{\linewidth}{l*{3}{>{\centering\arraybackslash}X}}
\toprule
\multicolumn{4}{c}{\textbf{DL3DV}} \\
\midrule
Source & PSNR$\uparrow$ & SSIM $\uparrow$ & F1$\uparrow$ \\
\midrule
Real event camera & -- & -- & -- \\
DVS-Voltmeter & 16.76 & 0.737 & 0.775 \\
v2e & 16.55 & 0.748 & 0.796 \\
Fixed-180 & 17.09 & 0.757 & 0.805 \\
Fixed-360 & \textbf{18.07} & \textbf{0.777} & \textbf{0.824} \\
\midrule
\rowcolor{ourgray}
EvGaze (Ours) & \underline{17.79} & \underline{0.762} & \underline{0.813} \\
\bottomrule
\end{tabularx}
\end{minipage}\hfill
\begin{minipage}[t]{0.53\textwidth}
\centering
\begin{tabularx}{\linewidth}{l*{4}{>{\centering\arraybackslash}X}}
\toprule
\multicolumn{5}{c}{\textbf{DSEC}} \\
\midrule
Source & PSNR$\uparrow$ & SSIM $\uparrow$ & F1$\uparrow$ & Far F1$\uparrow$ \\
\midrule
Real event camera & 14.35 & 0.606 & 0.795 & 0.425 \\
DVS-Voltmeter & 14.89 & 0.653 & 0.832 & 0.592 \\
v2e & 15.54 & 0.638 & 0.853 & 0.685 \\
Fixed-180 & 15.94 & \textbf{0.683} & \textbf{0.881} & 0.816 \\
Fixed-360 & \textbf{16.12} & 0.672 & \underline{0.880} & \textbf{0.823} \\
\midrule
\rowcolor{ourgray}
EvGaze (Ours) & \underline{16.11} & \underline{0.675} & \textbf{0.881} & \underline{0.820} \\
\bottomrule
\end{tabularx}
\end{minipage}
\end{table}

\paragraph{Event-Only 3D Reconstruction.}
We further investigate whether optically excited events improve 3D reconstruction quality on subsets of DL3DV and DSEC. We first reconstruct intensity frames from event streams using E2VID, followed by 3D Gaussian Splatting (3DGS)~\citep{kerbl20233dgs} optimization across three DL3DV scenes and three DSEC sequences. All methods share identical camera poses and COLMAP~\citep{schonberger2016structure} geometry initialization. As shown in \cref{tab:e2vid_3dgs}, EvGaze outperforms DVS-Voltmeter by 1.03 dB on DL3DV and 1.22 dB on DSEC in PSNR. It approaches the reconstruction quality of Fixed-360, with PSNR differences of only 0.28 dB and 0.01 dB, respectively, while achieving a joint-best edge F1 of 0.881 on DSEC. Complementary event-assisted reconstruction results from motion-blurred RGB following EvaGaussians~\citep{yu2025evagaussians} are provided in \cref{supp:evagaussians_method} (\cref{tab:supp_evagaussians_dl3dv,tab:supp_evagaussians_dsec}), where active excitation likewise improves over native events and passive simulation on DSEC without continuous full-speed scanning.

\subsection{Real-World Evaluation}
\label{sec:real_world_evaluation}
\label{sec:reconstruction_hardware}

\begin{table}[t]
\centering
\caption{\textbf{Real-world optical excitation and tracking.}
Edge F1, event rate, and feature tracking on our prototype.
Best results in bold, second-best underlined.}
\label{tab:real_combined}
\small
\setlength{\tabcolsep}{6pt}
\renewcommand{\arraystretch}{1.12}
\resizebox{0.7\linewidth}{!}{%
\begin{tabular}{lcccccc}
\toprule
\multirow{2}{*}{Source}
& \multicolumn{3}{c}{Edge F1$\uparrow$ / Rate (Mev/s)}
& \multicolumn{3}{c}{Tracking} \\
\cmidrule(lr){2-4}\cmidrule(lr){5-7}
& Static & Car & Ground
& S@1$\uparrow$ & S@3$\uparrow$ & Med.$\downarrow$ \\
\midrule
Fixed-0
& 0.133 / 0.01
& 0.583 / 0.14
& 0.567 / 0.11
& 0.266 & 0.452 & 4.69 \\
Fixed-180
& 0.924 / 0.81
& 0.907 / 0.96
& \textbf{0.917} / 0.84
& 0.759 & 0.973 & \underline{0.81} \\
Fixed-360
& \underline{0.941} / 1.74
& \underline{0.923} / 1.71
& \underline{0.902} / 1.64
& \textbf{0.867}
& \textbf{0.991}
& \textbf{0.60} \\
\midrule
EvGaze (ZS)
& 0.924 / 0.81
& 0.907 / 0.96
& 0.901 / 0.84
& 0.759 & 0.973 & \underline{0.81} \\
\rowcolor{ourgray}
EvGaze (FT)
& \textbf{0.954} / 1.37
& \textbf{0.938} / 1.31
& \textbf{0.917} / 1.53
& \underline{0.834}
& \underline{0.987}
& \textbf{0.60} \\
\bottomrule
\end{tabular}%
}
\end{table}

\begin{table}[t]
\centering
\caption{\textbf{Controller ablation and rule-based baselines.}
Average performance over four motion regimes and per-regime
$\Delta$F1 relative to EvGaze.
Best results in bold, second-best underlined.}
\label{tab:controller_ablation}
\small
\setlength{\tabcolsep}{6pt}
\renewcommand{\arraystretch}{1.12}
\resizebox{0.7\linewidth}{!}{%
\begin{tabular}{lccccccc}
\toprule
\multirow{2}{*}{Controller}
& \multicolumn{3}{c}{Average}
& \multicolumn{4}{c}{$\Delta$F1 by Motion Regime} \\
\cmidrule(lr){2-4}\cmidrule(lr){5-8}
& F1$\uparrow$ & TEO$\uparrow$ & Mev/s$\downarrow$
& Cam. & Co. & DSD & Obj. \\
\midrule
\rowcolor{ourgray}
EvGaze (Ours)
& \textbf{0.553} & \textbf{0.466} & \textbf{59.7}
& \textbf{0.000} & \textbf{0.000}
& \textbf{0.000} & \textbf{0.000} \\
\hspace{0.5em}w/o fast path
& 0.544 & \underline{0.455} & \underline{63.5}
& -0.006 & -0.024 & -0.007 & \underline{-0.002} \\
\hspace{0.5em}w/o IMU
& \underline{0.545} & 0.454 & 69.2
& \underline{-0.005} & -0.021
& \underline{-0.003} & -0.004 \\
\midrule
Rule: event activity
& 0.509 & 0.448 & 83.9
& -0.013 & \underline{-0.015} & -0.034 & -0.115 \\
Rule: IMU motion
& 0.505 & 0.445 & 93.1
& -0.020 & -0.022 & -0.034 & -0.115 \\
\bottomrule
\end{tabular}%
}
\end{table}

\paragraph{Physical Prototype Evaluation.}
We evaluate optical excitation and adaptive control on our hardware prototype under three conditions: a static scene, an object on a car, and an object on the ground (\cref{tab:real_combined}). In the static scene, rotating the prism at 360 RPM raises Edge F1 from 0.133 to 0.941, demonstrating the effectiveness of physical optical excitation. Fixed-360 also achieves the highest static-scene tracking success, consistent with its denser event sampling. While the zero-shot policy operates similarly to Fixed-180 due to the event-rate scale shift between synthetic rendering and physical acquisition, real-device fine-tuning enables EvGaze to achieve the highest Edge F1 across all three conditions, with 21\%--23\% lower event rates than Fixed-360 in the static and car-mounted scenarios. Overall, EvGaze maintains strong edge observability and feature tracking at reduced event rates.

\subsection{Ablation Studies}
\label{sec:component_ablation}

\paragraph{Controller and Inertial Feedback.}
We investigate the contributions of the learned controller and its inertial components (\cref{tab:controller_ablation}). Removing either the fast pathway or IMU conditioning reduces structural observability, with the largest degradation under co-motion. Rule-based controllers exhibit larger performance losses, particularly in the object-only regime, despite producing more events. These results highlight the importance of motion-aware control over simply increasing event activity.

\paragraph{Additional Analysis.}
We further investigate timestamp-aware optical compensation and event-accumulation fidelity. We compare uncompensated events, interval-wise compensation, and per-event alignment to examine the effect of temporal precision on motion correction. Per-event alignment improves edge quality and IWE sharpness at both tested rotation speeds, demonstrating the importance of accurate prism-motion compensation. Beyond edge-based evaluation, we employ FSIM~\citep{zhang2011fsim} and LPIPS~\citep{zhang2018perceptual} to assess the structural and perceptual fidelity of aligned event accumulations. Quantitative results for both analyses (\cref{tab:compensation_ablation,tab:accumulation_fidelity}) and measured motor step responses (\cref{fig:supp_imu_ablation}) are reported in \cref{supp:actuator,supp:analysis}.

\section{Conclusion}
\label{sec:conclusion}

We presented \textbf{EvGaze}, an adaptive event-sensing system that regulates optical excitation before events are generated. By controlling a variable-speed rotating wedge prism, EvGaze reveals otherwise unobserved structure under weak or single-axis motion while limiting redundant events once natural motion is sufficient. A learned multimodal slow pathway paired with a high-rate inertial fast pathway adapts prism speed under physical actuator constraints, while encoder-based timestamp-aware compensation preserves geometric alignment across speed transitions. Experiments on EvMotion, DSEC, and DL3DV demonstrate improved structural observability and favorable quality--bandwidth trade-offs that benefit downstream feature tracking and event-only 3D reconstruction, and real-device deployment validates online closed-loop adaptation on physical hardware.

\paragraph{Limitations and Future Work.}
Several limitations suggest promising directions for future work. First, rotary motor dead time $t_d$ and mechanical inertia $\tau$ restrict instantaneous speed transitions and motivate discrete adjacent-gear filtering; replacing the rotary drive with low-inertia beam-steering optics (e.g., piezo fast steering mirrors or electrowetting prisms) and synthesizing directional micro-saccades orthogonal to dominant ego-motion could further cut redundant events. Second, the current slow pathway incorporates external RGB luminance cues and relies on domain-matched fine-tuning to bridge the event-rate scale gap between synthetic rendering and physical sensors; scale-invariant pure Event--IMU control with self-supervised test-time adaptation would strengthen deployment in low-light and high-dynamic-range environments. Finally, coupling pre-event optical excitation with in-sensor bias regulation and downstream task uncertainty offers a natural path toward end-to-end task-driven active event sensing.

\clearpage

\bibliographystyle{plainnat}
\setlength{\bibhang}{0pt}
\setlength\bibindent{0pt}
{\setlength{\bibsep}{3.2pt}\bibliography{main}}

\clearpage
\beginappendix
\crefalias{section}{appendix}
\crefalias{subsection}{appendix}

\noindent\textbf{Appendix Overview.}
We present the optical geometry and event compensation of \textbf{EvGaze}, followed by the calibrated actuator model, dual-rate policy, and training procedure. We then detail the datasets, evaluation protocols, optical and inertial ablations, 3D reconstruction experiments, and physical-prototype implementation.

\section{Optical Formation, Inverse Mapping, and Calibration}
\label{supp:optics}
\label{app:optics}

\subsection{Single-wedge optical excitation}
A thin wedge prism deviates an incident ray by approximately $\delta\simeq(n-1)\alpha$, where $\alpha$ is the wedge apex angle and $n$ its refractive index. Rotation about the optical axis preserves the first-order deflection magnitude while rotating its direction. A single motor produces phase-dependent image-plane motion across edge orientations. The prism phase determines the inverse mapping for event alignment.

\begin{figure}[htbp]
\centering
\includegraphics[width=0.82\linewidth]{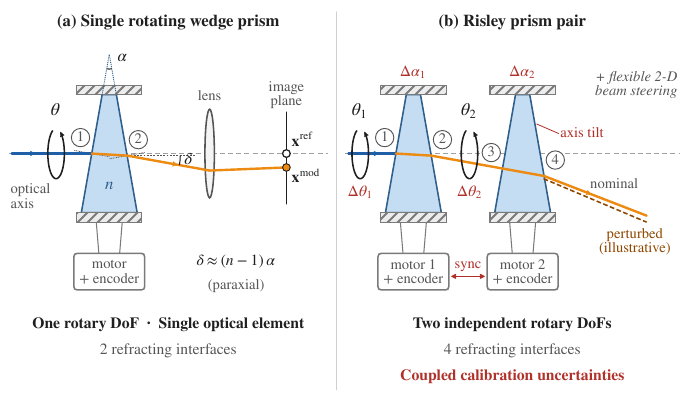}
\caption{\textbf{Optical design rationale for EvGaze.}
(a) A single rotating wedge prism provides phase-dependent optical deflection with one rotary degree of freedom.
(b) A Risley pair enables more flexible beam steering through two independently rotating wedges, with additional refracting interfaces and coupled calibration variables.}
\label{fig:supp_optical_design}
\end{figure}

A Risley pair uses two independently rotating wedges with four refracting interfaces, introducing coupled calibration variables across apex-angle tolerances, axial alignment, and relative phase synchronization~\citep{zhou2014risley_error,zhao2014risley_firstorder}. For EvGaze, a single wedge provides the required directional sweep with one actuator and a compact phase-to-image mapping. \cref{fig:supp_optical_design} compares the optical configurations and their calibration requirements.

\subsection{Forward optical model and induced velocity}
Let $\mathbf K$ be the intrinsic matrix and $\bar{\mathbf x}=[u,v,1]^\top$ a homogeneous reference pixel. We define the normalized camera coordinate $\mathbf p=\mathbf K^{-1}\bar{\mathbf x}$, $\rho(\mathbf x)=\|\mathbf p\|_2$, and $\mathbf b(\theta)=\delta[\cos\theta,\sin\theta]^\top$. The forward warp in \cref{eq:forward_warp} is
\begin{equation}
W_{\mathrm{opt}}(\mathbf x;\theta)=\mathbf x+
\rho(\mathbf x)\delta
\begin{bmatrix}f_x\cos\theta\\f_y\sin\theta\end{bmatrix}.
\label{eq:supp_forward}
\end{equation}
In normalized image coordinates, $\mathbf y=\mathbf p_{1:2}+\rho(\mathbf x)\mathbf b(\theta)$. For fixed $\rho$ and equal focal lengths, varying $\theta$ traces a circle; unequal focal lengths yield an ellipse in pixel coordinates. The modulated irradiance is formed \emph{before} event thresholding as $S(\mathbf x,t)=I(W_{\mathrm{opt}}^{-1}(\mathbf x;\theta(t)),t)$.

Differentiating \cref{eq:supp_forward} at a fixed reference coordinate yields the induced optical velocity:
\begin{equation}
\mathbf v_{\mathrm{opt}}(\mathbf x,t)=
\rho(\mathbf x)\delta\dot\theta(t)
\begin{bmatrix}
-f_x\sin\theta(t)\\
\phantom{-}f_y\cos\theta(t)
\end{bmatrix}.
\label{eq:supp_optical_velocity}
\end{equation}
For an image edge with local log-intensity gradient $\nabla\log I$ and natural apparent velocity $\mathbf v_{\mathrm{img}}$, the first-order brightness change is
\begin{equation}
\frac{\partial\log I}{\partial t}\simeq
-\nabla\log I\cdot
\bigl(\mathbf v_{\mathrm{img}}+\mathbf v_{\mathrm{opt}}\bigr).
\label{eq:supp_brightness_motion}
\end{equation}
Prism rotation injects an edge-normal velocity component whose direction sweeps with $\theta(t)$ and whose magnitude scales with $\dot\theta(t)$. Over a $16.7$\,ms window, constant $180$ and $360$\,RPM sweep approximately $18^\circ$ and $36^\circ$, respectively.

\subsection{Closed-form inverse and timestamp-aware alignment}
Given normalized sensor coordinate $\mathbf y$ at phase $\theta$, let $r=\rho(\mathbf x)$ and $\mathbf b=\mathbf b(\theta)$. From $\mathbf p_{1:2}=\mathbf y-r\mathbf b$ and $r^2=1+\|\mathbf p_{1:2}\|_2^2$, the inverse reduces to
\begin{equation}
(1-\|\mathbf b\|_2^2)r^2+2(\mathbf y^\top\mathbf b)r
-(1+\|\mathbf y\|_2^2)=0.
\label{eq:supp_quad}
\end{equation}
For $\|\mathbf b\|_2<1$, the physically valid positive root is
\begin{equation}
\begin{split}
r={}&\frac{-\mathbf y^\top\mathbf b+\sqrt{D}}{1-\|\mathbf b\|_2^2},\\
D={}&(\mathbf y^\top\mathbf b)^2+
(1-\|\mathbf b\|_2^2)(1+\|\mathbf y\|_2^2).
\end{split}
\label{eq:supp_quad_root}
\end{equation}
The reference ray follows from $\mathbf p_{1:2}=\mathbf y-r\mathbf b$ and projects via $\mathbf K$. For each modulated event $e_k^{\mathrm{mod}}=(\mathbf x_k^{\mathrm{mod}},t_k,p_k)$, alignment evaluates the inverse at the event timestamp:
\begin{equation}
\mathbf x_k^{\mathrm{ref}}=
W_{\mathrm{opt}}^{-1}\!\left(\mathbf x_k^{\mathrm{mod}};\theta(t_k)\right).
\label{eq:supp_event_comp}
\end{equation}
Event timestamps and polarities are retained after inverse mapping to reference coordinates.

In simulation, pre-event modulation uses pixel-center subpixel interpolation on the inverse-warped coordinates. Pixels mapped outside the source image are excluded using the corresponding validity mask. The GPU implementation processes event coordinates in floating point prior to downstream rasterization, evaluating $\theta(t_k)$ from the integrated realized angular velocity.

\subsection{Physical prototype and optical calibration}
\begin{figure}[!t]
    \centering
    \begin{tikzpicture}
        \node[anchor=south west, inner sep=0] (img) at (0,0) {%
            \includegraphics[width=0.72\linewidth, height=0.33\textheight, keepaspectratio]{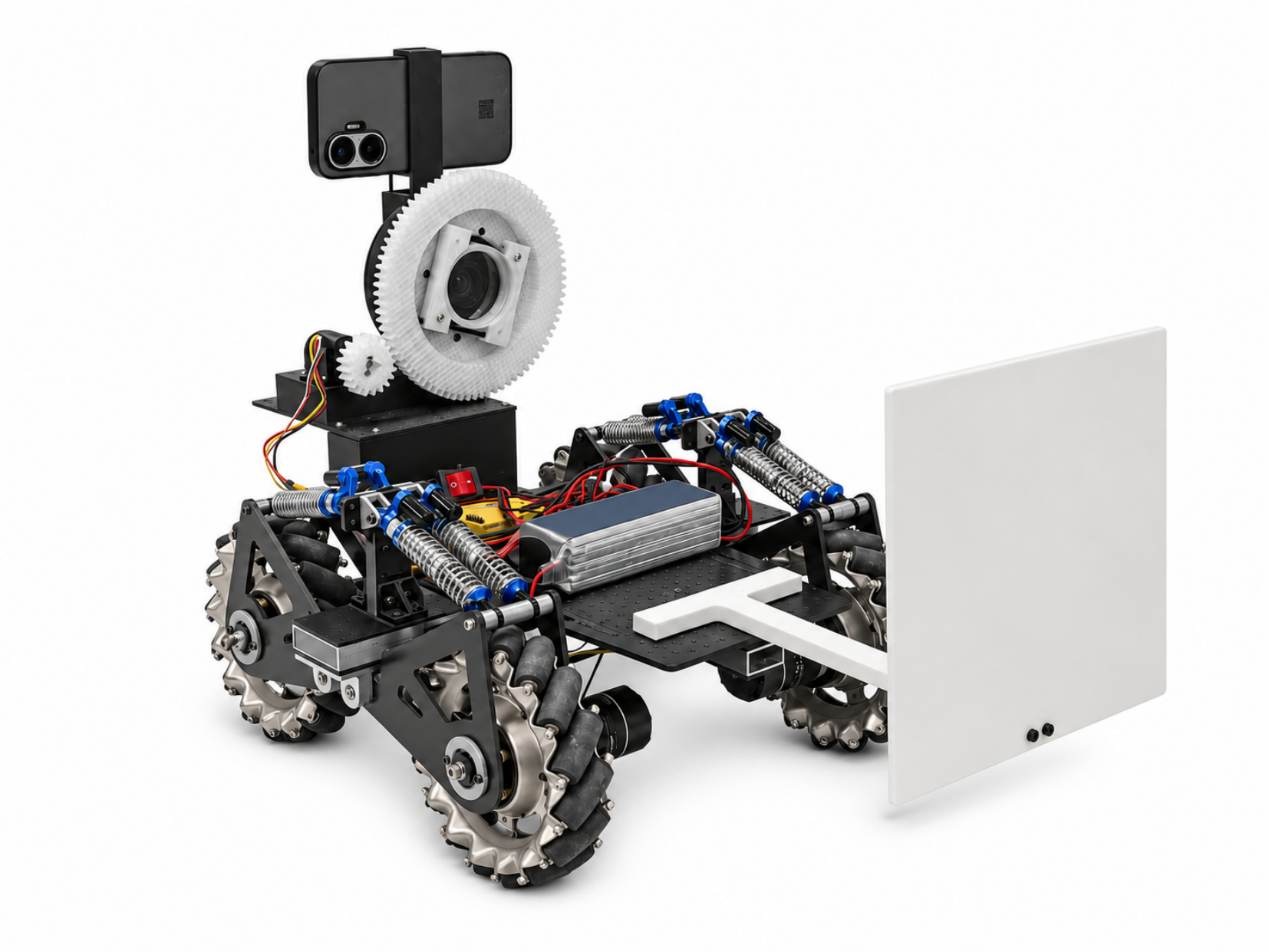}%
        };
        \begin{scope}[x=(img.south east), y=(img.north west),
            callout/.style={font=\sffamily\scriptsize\bfseries, text=black!85, fill=white, fill opacity=0.92, text opacity=1, rounded corners=1.5pt, inner sep=2pt, draw=black!30, line width=0.3pt},
            leader/.style={-stealth, line width=0.65pt, draw=pkured!85!black}]
            \node[callout, anchor=east] (phone) at (0.20, 0.88) {Smartphone RGB Camera};
            \draw[leader] (phone.east) -- (0.255, 0.855);

            \node[callout, anchor=east] (evcam) at (0.20, 0.78) {SilkyEvCam (behind prism)};
            \draw[leader] (evcam.east) -- (0.305, 0.785);

            \node[callout, anchor=east] (motor) at (0.20, 0.625) {M2006 Motor \& Encoder};
            \draw[leader] (motor.east) -- (0.255, 0.625);

            \node[callout, anchor=east] (base) at (0.20, 0.22) {Mecanum Mobile Chassis};
            \draw[leader] (base.east) -- (0.275, 0.22);

            \node[callout, anchor=west] (prism) at (0.49, 0.76) {Rotating Wedge \& 4:1 Gear};
            \draw[leader] (prism.west) -- (0.425, 0.73);

            \node[callout, anchor=west] (mcu) at (0.49, 0.63) {Damiao H7 MCU \& IMU};
            \draw[leader] (mcu.west) -- (0.382, 0.468);

            \node[callout, anchor=south] (panel) at (0.81, 0.70) {Front Target Panel};
            \draw[leader] (panel.south) -- (0.81, 0.60);
        \end{scope}
    \end{tikzpicture}
    \caption{\textbf{EvGaze physical prototype.}
    An M2006 motor drives the rotating wedge-prism assembly through a $4{:}1$ geared mechanism in front of the SilkyEvCam event camera. The Damiao H7 controller records inertial measurements and prism encoder angles, a smartphone camera captures external RGB observations, and the mecanum-wheeled chassis carries a front panel for car-mounted target evaluation.}
    \label{fig:supp_device}
\end{figure}

Our prototype uses a CenturyArks SilkyEvCam VGA event camera with a Prophesee PPS3MVCD sensor ($640\times480$). A DJI RoboMaster M2006 motor drives a rotating wedge assembly through a geared transmission in the camera optical path. The Damiao H7 lower-level controller records prism encoder angles and inertial measurements; a smartphone supplies the external RGB stream. \cref{fig:supp_device} shows the assembled mobile prototype and its major components.

The lower-level controller associates each encoder angle with an MCU timestamp. The host converts event-camera timestamps to the MCU time base and interpolates the unwrapped $100$\,Hz encoder-angle samples to the time of each event. The physical event inverse is a pixel shift opposite to the measured prism deflection:
\begin{equation}
\mathbf x_k^{\mathrm{ref}}=
\mathbf x_k^{\mathrm{mod}}-
\mathbf d_{\mathrm{hw}}\!\left(\theta_m(t_k^{\mathrm{MCU}})\right).
\label{eq:supp_hardware_inverse}
\end{equation}
The phase-dependent image displacement is expressed using the calibration model in \cref{sec:optical_physics},
\begin{equation}
\mathbf d_{\mathrm{hw}}(\theta_m)=R
\begin{bmatrix}
\cos(\phi_{\mathrm{hw}}-\theta_m)\\
\sin(\phi_{\mathrm{hw}}-\theta_m)
\end{bmatrix},
\label{eq:supp_hw_motion}
\end{equation}
where $R$ is the displacement magnitude in pixels and $\phi_{\mathrm{hw}}$ is the fitted phase offset. The pixel-shift scale depends on prism deflection and camera focal length. The displacement radius and phase offset are estimated by maximizing the contrast of compensated event accumulations from static scene structure~\citep{gallego2018contrast}. With $H_{R,\phi}$ denoting an accumulated image of inverse-shifted events, the objective is
\begin{equation}
(R^\star,\phi^\star)=\arg\max_{R,\phi}
\operatorname{Var}_{\mathbf x}\!\left[H_{R,\phi}(\mathbf x)\right].
\label{eq:supp_calibration_objective}
\end{equation}
Simulation uses the analytical warp in \cref{eq:supp_forward} with $\delta=0.5^\circ$. The physical compensation applies \cref{eq:supp_hardware_inverse} using the encoder angle at each event timestamp. Details of the clock alignment and data acquisition are given in \cref{supp:hardware_clock}.

\section{Actuator Dynamics and Command Execution}
\label{supp:actuator}

\subsection{Actuator Calibration}

The prism drive operates at $\mathcal A_p=\{0,180,360\}$ RPM. Its physical response is characterized by the first-order system with command delay in \cref{eq:motor_dynamics}:
\begin{equation}
\tau\dot s(t)=s_{\mathrm{cmd}}(t-t_d)-s(t),\qquad
\theta(t)=\theta_0+\frac{2\pi}{60}\int_0^t s(\xi)\,d\xi.
\label{eq:supp_motor}
\end{equation}

We calibrate the actuator dynamics on the physical prototype by measuring motor step responses across all four adjacent gear transitions. Motor-speed measurements are converted to prism RPM using the $4{:}1$ transmission ratio. The actuator model adopts the median calibrated command delay $t_d=0.1215$\,s and time constant $\tau=0.1113$\,s.

\subsection{Command Execution}

The controller converts action logits $\boldsymbol\ell(t)$ into probabilities $\mathbf p(t)=
\operatorname{softmax}(\boldsymbol\ell(t))$. A gear change is accepted when the candidate probability exceeds that of the current commanded gear by the hysteresis margin $\delta_p=0.1$ and at least $\Delta t_{\min}=0.4553$\,s has elapsed since the previous command. Transitions follow the adjacent-gear sequence $0\leftrightarrow180\leftrightarrow360$ RPM, routing an extreme-to-extreme request through $180$ RPM.

\section{Dual-Rate Controller and Policy Learning}
\label{supp:controller}
\label{app:controller}
\subsection{Multimodal inputs and temporal encoding}
\label{supp:token_construction}
The slow controller processes log-luminance variation, IMU motion, compensated-event occupancy, and event-support overlap. Their raw feature dimensions are $4$, $9$, $17$, and $17$, respectively (\cref{tab:supp_input_tokens}). Luminance features are the median, $90$th percentile, $99$th percentile, and mean absolute log-intensity variation per unit time. IMU features comprise three rotation increments, three angular velocities, and three linear accelerations. Occupancy contains a $4\times4$ spatial grid and a global occupancy fraction; overlap uses the corresponding intersection-over-union of successive compensated event supports. The controller uses a $0.5$\,s history on a $5$\,ms grid. Each modality is resampled with the most recent available observation and its age. With relative history time $\tau_j\le0$ and $H=0.5$\,s, the temporal encoding is
\begin{equation}
\mathbf e_t(\tau_j)=\left[
\frac{\tau_j}{H},\left\{
\sin\frac{2\pi\tau_j}{p},
\cos\frac{2\pi\tau_j}{p}
\right\}_{p\in\mathcal P}\right],
\label{eq:supp_time_features}
\end{equation}
where $\mathcal P=\{0.01,0.033,0.1,0.3,1.0\}$\,s. Training-set statistics normalize the input features before token projection.

\begin{table}[htbp]
\centering
\caption{\textbf{Multimodal input representation.} All event statistics are computed after prism-motion compensation.}
\label{tab:supp_input_tokens}
\small
\setlength{\tabcolsep}{6pt}
\renewcommand{\arraystretch}{1.12}
\begin{tabular*}{0.88\linewidth}{@{\extracolsep{\fill}}lcl@{}}
\toprule
Modality & Dim. & Input statistics\\
\midrule
Luminance & 4 & Median, $q_{90}$, $q_{99}$, mean $|\Delta\log I|/\Delta t$\\
Inertial & 9 & $\Delta\boldsymbol\theta$, $\bar{\boldsymbol\omega}$, $\bar{\mathbf a}$\\
Occupancy & 17 & $4\times4$ cell fractions + global fraction\\
Overlap & 17 & $4\times4$ cell IoU + global IoU\\
\bottomrule
\end{tabular*}
\end{table}

\subsection{Causal temporal encoder and slow pathway}
\label{supp:encoder_fusion}
Each modality is projected into $d=32$ channels and processed by six residual causal convolutions with kernel size $3$ and dilations $1,2,4,8,16,32$. The receptive field spans $127$ grid positions. The modality-specific token projection is
\begin{equation}
\mathbf h^{(0)}_{v,j}=\mathbf W_v
[\mathbf x_{v,j};\ a_{v,j};\ \mathbf e_t(\tau_j)]
+\mathbf b_v+\mathbf u_v,
\label{eq:supp_token_lift}
\end{equation}
where $a_{v,j}$ denotes observation age and $\mathbf u_v$ is a learned modality embedding. The final modality tokens are fused by four-head attention and LayerNorm alongside a context token containing $\sin\theta$, $\cos\theta$, realized RPM, commanded gear, actuator settling state, and bandwidth price $\lambda$. Concatenated attended modality tokens form a cached representation $\mathbf z\in\mathbb R^{128}$. The slow decision head outputs three logits at $60$\,Hz. \cref{tab:supp_controller_arch} summarizes the network configuration.

\begin{table}[htbp]
\centering
\caption{\textbf{Dual-rate controller architecture.} Slow temporal encoder and inertial residual branch.}
\label{tab:supp_controller_arch}
\small
\setlength{\tabcolsep}{6pt}
\renewcommand{\arraystretch}{1.12}
\begin{tabularx}{0.88\linewidth}{l*{2}{>{\centering\arraybackslash}X}}
\toprule
Component & Configuration & Parameters \\
\midrule
Input raw dimensions & 4 / 9 / 17 / 17 & -- \\
Temporal grid & 100 steps at 5\,ms & -- \\
Per-variable latent & $d=32$ & -- \\
Temporal dilations & $1,2,4,8,16,32$ & -- \\
Cross-modal attention & 4 heads, 1 layer & -- \\
Slow decision head & 136--64--64--3 & 97,027$^\dagger$ \\
Fast residual head & 201--64--64--3 & 28,867$^\dagger$ \\
\midrule
\rowcolor{ourgray}
Combined policy & -- & 125,894 \\
\bottomrule
\end{tabularx}
\par\vspace{2pt}{\scriptsize $^\dagger$Parameter counts include the complete modules.}
\end{table}

\subsection{Inertial fast pathway}
\label{supp:fast_details}
Between visual updates, the inertial residual refines the logits using cached multimodal features, updated inertial history, and actuator context:
\begin{equation}
\boldsymbol\ell(t)=\operatorname{head}_{\rm slow}(\mathbf z,\mathbf c(t))+
 g\!\left(\mathbf z,\mathbf u_{\rm imu}(t),\mathbf c(t),\operatorname{age}(t)\right).
\label{eq:supp_fast}
\end{equation}
The residual uses the latest $20\times9$ inertial features over a $100$\,ms interval, projects them to $64$ dimensions, and concatenates them with the $128$-dimensional cache, actuator context, and cache age. Its MLP has input width $201$, two $64$-unit hidden layers, and three output logits. Simulation evaluates the fast pathway on nominal $5$\,ms ($200$\,Hz) control steps. The physical controller receives timestamp-aligned IMU and encoder samples recorded at $100$\,Hz. Accepted speed commands pass through the filter in \cref{supp:actuator}.

\subsection{Actuator-aware dynamic-programming teacher}
\label{supp:teacher_dynamics}
The teacher evaluates each candidate optical action using the quality--bandwidth objective of \cref{eq:rate_distortion}:
\begin{equation}
J_\lambda(w,g)=Q(w,g)-\lambda\frac{r(w,g)}{r_{\rm ref}}.
\label{eq:supp_utility}
\end{equation}
Here $Q=(F_1+\mathrm{TEO})/2$ for target-observability experiments and $Q=F_1$ for whole-image experiments; $r(w,g)$ is the event rate of candidate gear $g$ in window $w$. The teacher uses matched fixed-speed optical rollouts and the delayed motor model in \cref{eq:supp_motor}. For actuator state $h_i$, admissible actions $\mathcal A(h_i)$, successor state $T_i(h_i,a)$, and transition utility $U_i(h_i,a)$, the finite-horizon recursion is
\begin{equation}
\begin{aligned}
V_i(h_i)&=\max_{a\in\mathcal A(h_i)} Q_i(h_i,a),\\
Q_i(h_i,a)&=U_i(h_i,a)+V_{i+1}(T_i(h_i,a)),\\
V_N(h_N)&=0.
\end{aligned}
\label{eq:supp_dp}
\end{equation}
Window utility integrates the interpolated gear utility along the realized speed trajectory $s_a(t)$:
\begin{equation}
U_i(h_i,a)=\int_{t_i}^{t_{i+1}}
\widetilde J_{\lambda,i}\bigl(s_a(t)\bigr)\,dt,
\label{eq:supp_integrated_reward}
\end{equation}
where $\widetilde J_{\lambda,i}$ interpolates the sampled action utilities as motor speed changes. During command delay, the response remains at the previous speed. The state transition enforces adjacent-gear moves and the minimum command interval. Action-wise regret is
\begin{equation}
R_\lambda(a\mid h_i)=\max_{a'\in\mathcal A(h_i)}Q_i(h_i,a')-Q_i(h_i,a).
\label{eq:supp_regret}
\end{equation}
Regret targets are computed for settled and transient actuator states.

\subsection{Policy optimization and training settings}
\label{supp:training_stages}
The policy is trained with expected regret, selective cross-entropy, and a gear-cycle penalty as in \cref{eq:policy_loss}:
\begin{equation}
\mathcal L=\mathbb E_{a\sim\pi_\psi}
\left[\frac{\min(R_\lambda(a),\kappa)}{\kappa}\right]
+\beta_{\rm CE}\mathcal L_{\rm CE}
+\beta_{\rm cyc}\mathcal L_{\rm cycle}.
\label{eq:supp_trainloss}
\end{equation}
We set $\kappa=0.25$, $\beta_{\rm CE}=0.2$, and $\beta_{\rm cyc}=1.0$. Cross-entropy is applied when the difference between the two smallest teacher regrets exceeds $0.005$. The cycle term penalizes immediate reversals in paired actuator states. The slow controller is trained for $40$ epochs with AdamW, learning rate $10^{-3}$, weight decay $10^{-4}$, cosine annealing, and batch size $256$. The bandwidth price is sampled from the training grid; the EvMotion evaluation (\cref{sec:experiments}) uses $\lambda=0.035$. A validation-regret criterion selects the checkpoint. The slow encoder and decision head are then frozen, and the inertial residual is trained for $10$ epochs. DSEC transfer uses leave-one-sequence-out fine-tuning at learning rate $3\times10^{-4}$. The hardware policy is adapted from recorded optical and inertial data before online deployment.

\section{Datasets and Event Generation}
\label{supp:data}
\subsection{EvMotion and motion regimes}
\label{supp:evmotion_regimes}
EvMotion consists of Blender sequences at $960\times540$ resolution and $60$\,fps, including Residential Avenue, Neon Auto District, and Cyberpunk Boulevard (\emph{CyberpunkCity}) (\cref{tab:supp_dataset_protocol}). Four motion regimes are evaluated: \emph{camera-only} (Cam.), with a moving camera and fixed target; \emph{object-only} (Obj.), with a fixed camera and moving target; \emph{co-motion} (Co.), with approximately fixed camera--target relative pose; and \emph{drive-stop-drive} (DSD), with motion separated by a stationary interval. The scenes provide synchronized RGB frames, camera trajectories, and projected target masks. RGB log-luminance defines reference edges; object masks specify TEO regions; camera trajectories supply synthetic inertial signals. All optical variants use the same source sequences, threshold, and scoring protocol. Controlled observability metrics are aggregated by regime over the common rendered duration. The feature-tracking evaluation uses $16.7$\,ms windows from these sequences.

\begin{table}[htbp]
\centering
\caption{\textbf{Datasets and principal evaluation settings.} Event-only reconstruction and motion-blur experiments use their respective scene subsets; tracking windows are listed separately from the controlled observability windows.}
\label{tab:supp_dataset_protocol}
\small
\setlength{\tabcolsep}{5pt}
\renewcommand{\arraystretch}{1.12}
\begin{tabular*}{0.92\linewidth}{@{\extracolsep{\fill}}lp{0.68\linewidth}@{}}
\toprule
Source & Dataset configuration\\
\midrule
EvMotion & Blender, $960\times540$, 60 fps; four relative-motion regimes (Cam., Obj., Co., DSD)\\
DSEC & Three driving sequences; registered RGB, recorded IMU and native-event reference\\
DL3DV & Reconstructed RGB scenes, RIFE temporal densification; scene subsets for qualitative evaluation and 3D reconstruction\\
Hardware & $640\times480$ event sensing; static, object-on-car, object-on-ground conditions\\
\bottomrule
\end{tabular*}
\end{table}

\subsection{DSEC and DL3DV}
DSEC~\citep{gehrig2021dsec} evaluation uses \texttt{thun\_00\_a}, \texttt{zurich\_city\_11\_a}, and the central six seconds of \texttt{zurich\_city\_03\_a}. RGB is registered to the event-camera image plane and temporally densified before prism modulation. The original event streams provide the native-camera reference. Tracking uses $159$ windows of $16.7$\,ms. Policy adaptation follows a leave-one-sequence-out procedure with recorded IMU data.

For DL3DV~\citep{ling2024dl3dv}, calibrated scene views are re-rendered and temporally densified with Practical-RIFE~\citep{huang2022rife}. The event-only reconstruction subset comprises scenes \texttt{03f5}, \texttt{114a}, and \texttt{1739}; input sequences at nominal $30$\,fps are densified to $480$\,fps. The reconstruction experiments use matched camera poses, timestamps, and held-out RGB views.

\subsection{Temporal densification and optical input}
\label{supp:temporal_densification}
For EvMotion, RGB observations and the scene trajectory are interpolated into intermediate brightness samples. DSEC and DL3DV RGB sequences are temporally densified with Practical-RIFE. At each intermediate time, the sensor image is formed as
\begin{equation}
 S(\mathbf x,t)=I\!\left(W_{\rm opt}^{-1}(\mathbf x;\theta(t)),t\right).
 \label{eq:supp_image_formation}
\end{equation}
Optical modulation is applied before event-threshold generation. DVS-Voltmeter~\citep{lin2022dvsvoltmeter} and v2e~\citep{hu2021v2e} process the matched RGB sources using their own event-generation backends.

\subsection{Stateful threshold-event generation}
\label{supp:event_simulator_details}
The simulator maintains a log-intensity reference $L_{\rm ref}$ at each pixel~\citep{rebecq2018esim,gallego2022event}, with $L(\mathbf x,t)=\log\max\{S(\mathbf x,t),\epsilon\}$. Positive and negative threshold crossings use symmetric contrast magnitude $C_{\rm th}=0.2$. A crossing of log-intensity level $L_c$ between $(t_0,L_0)$ and $(t_1,L_1)$ receives timestamp
\begin{equation}
 t_e=t_0+\frac{L_c-L_0}{L_1-L_0}(t_1-t_0),
\label{eq:supp_threshold_time}
\end{equation}
The reference level is updated after each crossing; multiple threshold crossings in one interval generate multiple events. Simulated optical phase advances by at most $1.5^\circ$ per sampling substep. At $360$ RPM, the phase advances $36^\circ$ over $1/60$\,s, corresponding to $24$ optical substeps at this spacing. Generated sensor events retain position, timestamp, and polarity, followed by the per-event inverse mapping in \cref{eq:supp_event_comp}. Fixed-0, Fixed-180, Fixed-360, and EvGaze use this simulation path with their respective prism trajectories.

\section{Evaluation Protocols}
\label{supp:metrics}
\subsection{Whole-image edge support and event rate}
\label{supp:edge_support}
Reference edges are obtained from RGB log-luminance gradient magnitudes using the $75$th percentile of nonzero values as the threshold. Let $G$ be the resulting binary reference-edge support and $M$ the set of pixels occupied by compensated events within the valid comparison region. Whole-image edge precision, coverage, and F1 are
\begin{equation}
\begin{aligned}
P_{\rm edge}&=\frac{|G\cap M|}{|M|},\qquad
C_{\rm edge}=\frac{|G\cap M|}{|G|},\\
F_1&=\frac{2P_{\rm edge}C_{\rm edge}}{P_{\rm edge}+C_{\rm edge}}.
\end{aligned}
\label{eq:supp_edge_f1}
\end{equation}
Each occupied pixel contributes once to $M$ regardless of polarity or the number of events at that pixel. Scores are set to zero for vanishing required denominators. In the DSEC edge-matching evaluation, both reference and event support use a one-pixel spatial tolerance. Event rate is computed from generated event counts divided by the acquisition interval and reported in Mev/s. Metrics are aggregated across their evaluation windows and sequences.

\subsection{Target Event Observability and persistence}
\label{supp:teo_temporal}
For target mask $T$, target reference edges $G\subseteq T$, and compensated event support $M$, the Target Event Observability (TEO) score uses target coverage $C$, target purity $P$, and persistence $\eta$:
\begin{equation}
\begin{aligned}
C&=\frac{|G\cap M|}{|G|},\quad
P=\frac{|G\cap M|}{|M\cap T|},\\
\mathrm{TEO}&=\frac{2CP}{C+P}\eta.
\end{aligned}
\label{eq:supp_teo}
\end{equation}
The evaluation window is partitioned into four equal time bins. Let $b_j$ indicate whether bin $j$ contains at least one aligned event supporting a reference target edge. Temporal persistence is
\begin{equation}
\eta=\frac{1}{4}\sum_{j=1}^{4} b_j.
\label{eq:supp_teo_persistence}
\end{equation}
TEO is zero for empty required support or $C+P=0$. Windows with empty reference target edges are excluded from its aggregation.

\subsection{Feature tracking and reference displacement}
\label{supp:tracking_details}
Each $16.7$\,ms window is divided into two event accumulations. Event counts are rasterized independently of polarity and log-compressed. Up to $400$ Shi--Tomasi~\citep{shi1994good} corners are detected in the first accumulation (quality level $0.01$, minimum distance $8$ pixels, block size $7$). Pyramidal Lucas--Kanade~\citep{lucas1981iterative} tracks them to the second using a $21\times21$ window and three pyramid levels. Tracks exiting the valid image region are excluded. For estimated displacement $\widehat{\mathbf d}_i$ and RGB-derived reference displacement $\mathbf d_i^{\rm ref}$, endpoint error and tracking success are
\begin{equation}
 e_i=\|\widehat{\mathbf d}_i-\mathbf d_i^{\rm ref}\|_2,
 \qquad \mathrm S@k=\frac{1}{N}\sum_{i=1}^{N}\mathbb I[e_i\le k].
 \label{eq:supp_tracking_success}
\end{equation}
We report $\mathrm S@1$, $\mathrm S@3$, and the median endpoint error. DSEC reference edges are stratified by displacement per $16.7$\,ms window: far-field ($<0.5$ px), intermediate ($0.5$--$2$ px), and fast ($\ge2$ px). Far-field edge coverage uses one-pixel spatial matching tolerance.

\subsection{Reconstruction and accumulation metrics}
\label{supp:structural_metrics}
Event-only E2VID~\citep{Rebecq19cvpr} + 3DGS~\citep{kerbl20233dgs} is evaluated against held-out RGB views using PSNR, SSIM~\citep{wang2004image}, and reconstructed-edge F1. Reconstructed renders are aligned to the reference brightness scale before evaluation. For accumulation fidelity, aligned event counts are compressed separately by polarity and combined into a normalized image-like representation. The reference is the normalized log-luminance gradient magnitude of the RGB image. FSIM~\citep{zhang2011fsim} and LPIPS~\citep{zhang2018perceptual} measure structural and perceptual agreement with this reference on the full EvMotion split and co-motion subset. Image-of-warped-events (IWE) sharpness~\citep{gallego2018contrast} is computed as the spatial variance of accumulated aligned event votes.

\section{Additional Analyses and Ablations}
\label{supp:analysis}
\label{app:ablations}
\subsection{Timestamp-aware optical compensation}
\label{supp:temporal_comp_reason}
The event stream is fixed while the compensation timing varies among no inverse mapping, interval-wise inverse mapping, and per-event inverse mapping (\cref{tab:compensation_ablation}). Per-event compensation raises Edge F1 from $0.5262$ to $0.5535$ at $180$ RPM and from $0.4875$ to $0.5288$ at $360$ RPM relative to interval-wise compensation. At $360$ RPM, the relative IWE sharpness reaches $\times1.473$. For interval-center phase $\theta(\bar t)$, the displacement difference of an event at $t_k$ is
\begin{equation}
\Delta\mathbf x_k=
W_{\rm opt}(\mathbf x_k;\theta(t_k))-
W_{\rm opt}(\mathbf x_k;\theta(\bar t)).
\label{eq:supp_interval_residual}
\end{equation}
For small $|t_k-\bar t|$, the leading-order residual scales with $\|\mathbf v_{\rm opt}(\mathbf x_k,\bar t)\|\,|t_k-\bar t|$. The faster phase sweep at $360$ RPM produces a larger within-window alignment residual.

\begin{table}[htbp]
\centering
\caption{\textbf{Ablation of optical motion compensation.}
Edge quality and relative IWE sharpness at 180 and 360 RPM.
Best values within each speed are bold.}
\label{tab:compensation_ablation}
\small
\setlength{\tabcolsep}{5pt}
\renewcommand{\arraystretch}{1.12}
\begin{tabularx}{0.88\linewidth}{ll*{4}{>{\centering\arraybackslash}X}}
\toprule
RPM & Compensation & F1$\uparrow$ & Prec.$\uparrow$ & Cov.$\uparrow$ & IWE$\uparrow$ \\
\midrule
180 & None & 0.4673 & 0.3219 & 0.8589 & $\times 1.000$ \\
    & Interval & 0.5262 & 0.3637 & 0.9579 & $\times 1.000$ \\
    & \cellcolor{ourgray}Per-event & \cellcolor{ourgray}\textbf{0.5535} & \cellcolor{ourgray}\textbf{0.3898} & \cellcolor{ourgray}\textbf{0.9649} & \cellcolor{ourgray}$\times\mathbf{1.180}$ \\
\midrule
360 & None & 0.4637 & 0.3082 & 0.9388 & $\times 1.000$ \\
    & Interval & 0.4875 & 0.3249 & 0.9789 & $\times 1.005$ \\
    & \cellcolor{ourgray}Per-event & \cellcolor{ourgray}\textbf{0.5288} & \cellcolor{ourgray}\textbf{0.3613} & \cellcolor{ourgray}\textbf{0.9915} & \cellcolor{ourgray}$\times\mathbf{1.473}$ \\
\bottomrule
\end{tabularx}
\end{table}

\subsection{Event-accumulation fidelity}
\cref{tab:accumulation_fidelity} reports FSIM and LPIPS for the full EvMotion test set and co-motion subset. EvGaze has the highest full-test FSIM ($0.6026$). Its full-test LPIPS is $0.4944$, compared with $0.4943$ for Fixed-180. On co-motion, EvGaze and Fixed-180 both achieve FSIM $0.6005$ and LPIPS $0.4870$.

\begin{table}[htbp]
\centering
\caption{\textbf{Event-accumulation fidelity on EvMotion.}
FSIM and LPIPS on the full test split and co-motion subset.
Best results are bold and second-best results are underlined.}
\label{tab:accumulation_fidelity}
\small
\setlength{\tabcolsep}{5pt}
\renewcommand{\arraystretch}{1.12}
\begin{tabularx}{0.88\linewidth}{l*{4}{>{\centering\arraybackslash}X}}
\toprule
 & \multicolumn{2}{c}{Full test} & \multicolumn{2}{c}{Co-motion} \\
\cmidrule(lr){2-3}\cmidrule(l){4-5}
Method & FSIM$\uparrow$ & LPIPS$\downarrow$ & FSIM$\uparrow$ & LPIPS$\downarrow$ \\
\midrule
DVS-Voltmeter & 0.5146 & 0.7396 & 0.5604 & 0.6622 \\
v2e & 0.5204 & 0.7105 & 0.5629 & 0.6118 \\
\midrule
Fixed-180 & \underline{0.6021} & \textbf{0.4943} & \textbf{0.6005} & \textbf{0.4870} \\
Fixed-360 & 0.5637 & 0.5102 & \underline{0.5696} & \underline{0.4983} \\
\midrule
\rowcolor{ourgray}
EvGaze (Ours) & \textbf{0.6026} & \underline{0.4944} & \textbf{0.6005} & \textbf{0.4870} \\
\bottomrule
\end{tabularx}
\end{table}

\subsection{Inertial-pathway and actuator analysis}
\begin{figure}[htbp]
\centering
\includegraphics[width=0.88\linewidth]{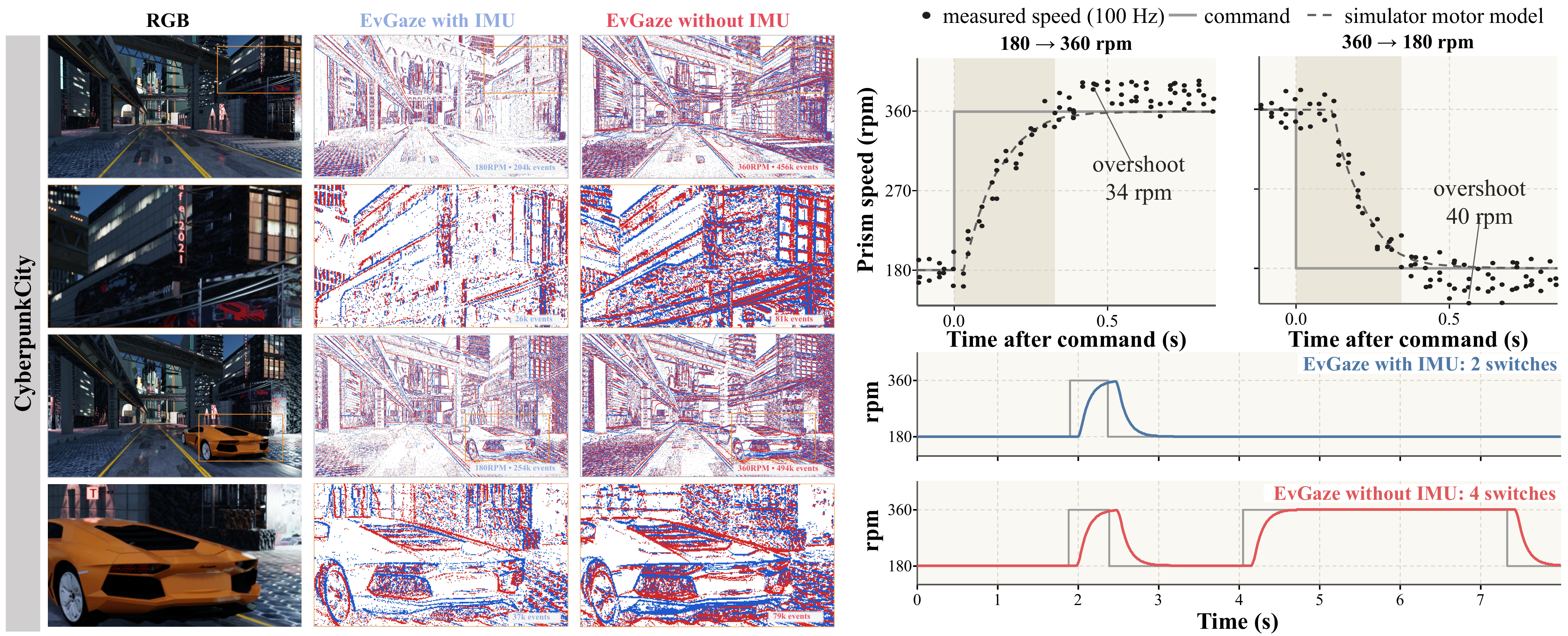}
\caption{\textbf{IMU ablation and actuator dynamics.} Inertial conditioning reduces background clutter on \emph{CyberpunkCity} (left). The lower-right sequence illustrates fewer commanded prism-speed switches, from four to two. Measured motor step responses are shown at the upper right.}
\label{fig:supp_imu_ablation}
\label{fig:imu_ablation}
\end{figure}

The controller ablation (\cref{tab:controller_ablation}) reports Edge F1/TEO of $0.553/0.466$ for EvGaze, $0.544/0.455$ after removing the fast pathway, and $0.545/0.454$ after removing IMU input. \cref{fig:supp_imu_ablation} shows the \emph{CyberpunkCity} event accumulations, measured speed responses to motor commands, and a sample sequence with two commanded switches for EvGaze with IMU versus four for its IMU ablation.

\section{Additional 3D Reconstruction Experiments}
\label{supp:reconstruction}
\subsection{Event-only E2VID + 3DGS}
\label{supp:eventonly_pipeline}
Aligned event streams are converted into reconstructed intensity frames by E2VID, using approximately $0.35WH$ events per update. A shared sparse COLMAP~\citep{schonberger2016structure} initialization and the same camera poses are used to fit 3D Gaussian Splatting models. Thirty frames are selected per scene, with $25$ training views and five held-out views. Each model is optimized for $7{,}000$ iterations. All event sources use the same selected view times, initialization, and training budget. \cref{tab:e2vid_3dgs} reports PSNR, SSIM, and reconstructed-edge metrics on DL3DV and DSEC.

\subsection{Event-assisted reconstruction with motion-blurred RGB}
\label{supp:evagaussians_method}
We use the CoherentGS-style exposure protocol~\citep{feng2026breaking} and the event-consistency objective of EvaGaussians~\citep{yu2025evagaussians}. Practical-RIFE provides dense RGB frames at $480$\,fps. Each blurred observation is the average of $48$ frames over a $0.1$\,s exposure:
\begin{equation}
B(\mathbf x)=\frac{1}{N}\sum_{j=1}^{N}I(\mathbf x,t_j),
\qquad t_j\in[t_a,t_b].
\label{eq:supp_blur_average}
\end{equation}
The Gaussian model is trained with $20$ blurred RGB views, nine virtual exposure poses per view, and $15{,}000$ iterations. The photometric term combines $L_1$ and SSIM; event consistency uses the change of rendered log intensity and the accumulated event polarities. Event-loss weight is $5\times10^{-3}$ after a $3{,}000$-iteration warm-up. The reconstructed sharp views are evaluated against matched held-out RGB references. \cref{tab:supp_evagaussians_dl3dv,tab:supp_evagaussians_dsec} report the results.

\begin{table}[htbp]
\centering
\caption{\textbf{Event-assisted 3D reconstruction on DL3DV.} Mean metrics over four DL3DV scenes under matched motion-blur and event-assisted training protocol. Best results in bold, second-best underlined.}
\label{tab:supp_evagaussians_dl3dv}
\small
\setlength{\tabcolsep}{5pt}
\renewcommand{\arraystretch}{1.12}
\begin{tabularx}{0.88\linewidth}{l*{3}{>{\centering\arraybackslash}X}}
\toprule
Event source & PSNR$\uparrow$ & SSIM$\uparrow$ & LPIPS$\downarrow$\\
\midrule
DVS-Voltmeter & 29.20 & 0.8696 & 0.2087\\
v2e & 29.23 & 0.8705 & \underline{0.2077}\\
\midrule
Fixed-0 & 29.20 & 0.8704 & \textbf{0.2076}\\
Fixed-180 & \textbf{29.26} & \textbf{0.8709} & 0.2078\\
Fixed-360 & \underline{29.25} & 0.8705 & 0.2078\\
\midrule
\rowcolor{ourgray}
EvGaze (Ours) & \textbf{29.26} & \underline{0.8707} & 0.2078\\
\bottomrule
\end{tabularx}
\end{table}

\begin{table}[htbp]
\centering
\caption{\textbf{Event-assisted 3D reconstruction on DSEC \texttt{thun\_00\_a}.} Sharp-view reconstruction after 15k optimization iterations, evaluated on 35 held-out views. Best results in bold, second-best underlined.}
\label{tab:supp_evagaussians_dsec}
\small
\setlength{\tabcolsep}{5pt}
\renewcommand{\arraystretch}{1.12}
\begin{tabularx}{0.88\linewidth}{l*{3}{>{\centering\arraybackslash}X}}
\toprule
Event source & PSNR$\uparrow$ & SSIM$\uparrow$ & LPIPS$\downarrow$\\
\midrule
Native events & 28.33 & 0.8670 & 0.2586\\
DVS-Voltmeter & 28.36 & 0.8681 & 0.2578\\
v2e & \textbf{28.50} & 0.8682 & 0.2545\\
\midrule
Fixed-180 & 28.44 & 0.8683 & \underline{0.2542}\\
Fixed-360 & \underline{28.49} & \textbf{0.8691} & \textbf{0.2524}\\
\midrule
\rowcolor{ourgray}
EvGaze (Ours) & 28.48 & \underline{0.8689} & 0.2543\\
\bottomrule
\end{tabularx}
\end{table}

Under this $0.1$\,s exposure setting, photometric supervision ($L_1$ and SSIM, with event-loss weight $5\times10^{-3}$) already anchors global scene appearance on DL3DV, so all event sources cluster within $29.20$--$29.26$\,dB PSNR. On DSEC \texttt{thun\_00\_a}, active optical excitation improves over native events ($28.33$\,dB) and DVS-Voltmeter ($28.36$\,dB), with EvGaze reaching $28.48$\,dB PSNR, $0.8689$ SSIM, and $0.2543$ LPIPS without continuous $360$\,RPM scanning.

\section{Physical Prototype and Online Closed-Loop Evaluation}
\label{supp:hardware}
\subsection{Hardware acquisition and sensing}
The physical prototype uses the $640\times480$ SilkyEvCam event sensor, rotating wedge, DJI M2006 drive with $4{:}1$ gearing, prism-angle encoder, Damiao H7 controller with IMU, and a smartphone RGB camera (\cref{fig:supp_device}). The MCU records IMU, encoder angle, motor feedback, and its millisecond counter at approximately $100$\,Hz. The event sensor acquires asynchronous events; RGB frames provide visual observations and registered image references.

\subsection{Clock calibration and event-phase synchronization}
\label{supp:hardware_clock}
The event camera, smartphone camera, and MCU maintain independent clocks. The MCU millisecond counter defines the common time axis. Fifteen screen flashes determine the event- and RGB-clock scales; one flash before each recording determines session-specific offsets. For sensor timestamp $t^{(j)}$ in milliseconds, the affine clock mapping is
\begin{equation}
 t_{\mathrm{MCU}}=a_j t^{(j)}+b_j,
 \qquad j\in\{\mathrm{ev},\mathrm{rgb}\}.
\label{eq:supp_clock_affine}
\end{equation}
For the event sensor's microsecond timestamps, conversion uses
\begin{equation}
 t_k^{\mathrm{MCU}}=a_{\mathrm{ev}}\frac{t_k^{\mathrm{ev}}}{1000}+b_{\mathrm{ev}}.
\label{eq:supp_event_clock}
\end{equation}
Flash times are located using event-count peaks in $1$\,ms bins and changes in RGB screen brightness. The calibrated clock scales are $a_{\rm ev}=0.99936$ and $a_{\rm rgb}=0.99884$. Encoder phase and IMU samples are recorded on the MCU clock. Unwrapped prism angle is linearly interpolated to each aligned event time:
\begin{equation}
 \theta_m(t_k^{\mathrm{MCU}})=\theta_i+
 \frac{t_k^{\mathrm{MCU}}-t_i}{t_{i+1}-t_i}
 (\theta_{i+1}-\theta_i).
\label{eq:supp_encoder_interpolation}
\end{equation}
Here $t_i\le t_k^{\rm MCU}<t_{i+1}$ and $\theta_i,\theta_{i+1}$ are consecutive unwrapped encoder measurements. The per-event hardware inverse warp uses the interpolated angle in \cref{eq:supp_hardware_inverse}. RGB frame timestamps are mapped to the MCU clock and the images are registered to event-camera coordinates.

\subsection{Online control and physical evaluation}
For online deployment, using the calibrated clock scales and per-session flash offset (\cref{supp:hardware_clock}), the host aligns incoming RGB, event, and inertial observations on the common MCU time base and commands $0$, $180$, or $360$ RPM through the hysteresis, adjacency, and dwell filter in \cref{supp:actuator}. The lower-level controller executes motor commands and returns realized speed and encoder phase. Event compensation uses the phase interpolated at each event timestamp during steady rotation and transitions.

Nine fixed-speed $15$\,s recordings spanning the three operating conditions supply optical calibration and policy adaptation. The deployed policy then controls prism speed online on the physical prototype. The three evaluation conditions are a static scene, an object on a car, and an object on the ground. \cref{tab:real_combined} reports Edge F1 and event rate, with static-scene feature tracking. EvGaze FT reaches $0.954$ Edge F1 at $1.37$\,Mev/s on the static scene, $0.938$ at $1.31$\,Mev/s for the car-mounted object, and $0.917$ at $1.53$\,Mev/s for the ground object. Continuous $360$ RPM yields $0.941$ at $1.74$\,Mev/s in the static scene and $0.923$ at $1.71$\,Mev/s on the car-mounted object.

\section{Implementation Details}
\label{supp:repro}
\subsection{Optical simulation and event alignment}
\label{supp:implementation_details}
The simulation pipeline executes image-sequence interpolation, phase-dependent optical resampling, per-pixel log-threshold event generation, and timestamp-aware inverse mapping. GPU tensor operations evaluate the analytical optical warp and compensate event coordinates in batches. Events retain sensor-space coordinates, timestamps, polarities, and floating-point reference-plane positions. Evaluation accumulations are rasterized from compensated positions. On physical hardware, encoder phase and the circular-translation calibration in \cref{eq:supp_hw_motion} determine the event-coordinate shift.

\subsection{Event Generation and Timestamp-Aware Alignment}
\label{app:evgaze_timestamp_alignment}

All fixed-speed and adaptive optical policies use the same temporal subdivision, calibrated prism model, and stateful log-intensity threshold generator. Optical modulation is applied to the scene luminance before event generation. Each threshold crossing produces an event with an interpolated timestamp, sensor-plane position, and polarity. For an event at time $t_e$, the corresponding prism phase $\theta(t_e)$ determines the inverse optical mapping to the reference plane. This per-event alignment removes prism-induced displacement while preserving natural camera and object motion, event timestamps, and polarities. The aligned event stream is used for downstream evaluation of optically modulated policies.

\subsection{CUDA-Accelerated Event Simulation and Runtime}
\label{app:evgaze_cuda_efficiency}

\paragraph{GPU implementation.}
The EvGaze optical simulation pipeline performs prism warping, threshold-event generation, and timestamp-aware compensation on the GPU. First, the calibrated inverse prism mapping identifies the source-image coordinates for each sensor pixel, and GPU tensor operations resample scene luminance with bilinear interpolation. Next, the stateful log-threshold generator processes each pixel's temporal samples in order. We implement this generator with CPU NumPy, PyTorch CUDA, and a compiled CUDA extension. The compiled extension uses two passes: the first counts threshold crossings per pixel, an exclusive prefix sum determines disjoint output offsets, and the second writes the events into an exactly sized buffer. The write pass also updates the per-pixel log-intensity reference for the next window. Finally, GPU-batched compensation evaluates the prism phase at each event timestamp and maps the resulting events to floating-point reference-plane coordinates. The closed-loop pipeline uses PyTorch CUDA by default and also supports the compiled event generator as an optional backend.

\paragraph{Numerical consistency.}
We evaluate the three implementations on 32 deterministic prism-warped frames at $960\times540$, using the same sample timestamps and contrast threshold $C=0.2$. \cref{tab:evgaze_cuda_consistency} reports identical event count and polarity statistics across all three backends. Event positions and polarities agree for every event. Relative to CPU NumPy, the maximum absolute differences are $1.11\times10^{-15}$\,s for event timestamps and $4.44\times10^{-16}$ for the final log-intensity reference. The two GPU implementations produce byte-identical, time-ordered event streams in this test.

\begin{table}[htbp]
\centering
\caption{\textbf{Numerical consistency across event-generation backends.}
Results on 32 deterministic prism-warped frames at
$960\times540$ with contrast threshold $C=0.2$.}
\label{tab:evgaze_cuda_consistency}
\small
\setlength{\tabcolsep}{5pt}
\renewcommand{\arraystretch}{1.12}
\begin{tabularx}{0.88\linewidth}{l*{3}{>{\centering\arraybackslash}X}}
    \toprule
    Backend & Total Events & Positive & Negative \\
    \midrule
    CPU NumPy     & 1,508,440 & 753,812 & 754,628 \\
    PyTorch CUDA  & 1,508,440 & 753,812 & 754,628 \\
    \rowcolor{ourgray}
    Compiled CUDA & 1,508,440 & 753,812 & 754,628 \\
    \bottomrule
\end{tabularx}
\end{table}

\paragraph{Event-generation throughput.}
We benchmark event generation on an NVIDIA GeForce RTX 4090 using 32 precomputed prism-warped luminance frames at $960\times540$ and 60\,Hz. Each simulator receives the matched frame sequence and timestamps, with its native input representation. Runtime includes event packing and ordering. We report the mean of three timed runs after one warm-up, excluding disk I/O, optical pre-warping, controller execution, and extension compilation. As shown in \cref{tab:evgaze_cuda_throughput}, PyTorch CUDA reaches 561.5 input FPS and the compiled CUDA extension reaches 1489.7 input FPS. Relative to the CPU NumPy implementation of the same threshold model, they achieve $15.63\times$ and $41.46\times$ acceleration, respectively. Compiled CUDA further improves event-generation throughput by $2.65\times$ over PyTorch CUDA.

\begin{table}[htbp]
\centering
\caption{\textbf{Event-generation throughput on RTX 4090.}
Measured on 32 matched input frames at $960\times540$.
Speedup is relative to EvGaze CPU NumPy. Best results in bold, second-best underlined.}
\label{tab:evgaze_cuda_throughput}
\small
\setlength{\tabcolsep}{5pt}
\renewcommand{\arraystretch}{1.12}
\begin{tabularx}{0.88\linewidth}{l*{3}{>{\centering\arraybackslash}X}}
    \toprule
    Method & Runtime (s)$\downarrow$ & FPS$\uparrow$ & Speedup$\uparrow$ \\
    \midrule
    DVS-Voltmeter (CPU) & 2.5874 & 12.4 & -- \\
    v2e (CPU)           & 0.3950 & 81.0 & -- \\
    EvGaze CPU NumPy    & 0.8907 & 35.9 & $1.00\times$ \\
    \midrule
    EvGaze PyTorch CUDA & \underline{0.0570} & \underline{561.5} & \underline{$15.63\times$} \\
    \rowcolor{ourgray}
    EvGaze Compiled CUDA
        & \textbf{0.0215}
        & \textbf{1489.7}
        & $\mathbf{41.46\times}$ \\
    \bottomrule
\end{tabularx}
\end{table}

\paragraph{Full optical-pipeline runtime.}
We also profile the full GPU optical pipeline on 32 rendered frames with 745 temporally subdivided samples. At 360\,RPM, the $1.5^\circ$ maximum phase increment yields 24 optical substeps per frame interval. The warp stage includes log-linear scene interpolation and valid-mask transfer; event generation includes packing and host transfer; and compensation includes per-event phase evaluation and reference-plane output transfer. \cref{tab:evgaze_cuda_pipeline} reports the mean stage times and an independently measured wall-clock total over three timed runs after one warm-up. The compiled generator reduces the generation stage from 0.753\,s to 0.125\,s ($6.01\times$) and the complete optical pipeline from 1.162\,s to 0.528\,s ($2.20\times$). Both pipelines generate 1,652,196 events. With compiled CUDA, the prism-warp and temporal-subdivision stage accounts for the largest share of the optical pipeline runtime.

\begin{table}[htbp]
\centering
\caption{\textbf{GPU optical-pipeline runtime on RTX 4090.}
Mean stage times for 32 frames and 745 optical samples.
Both backends generate 1,652,196 events. Best total in bold.}
\label{tab:evgaze_cuda_pipeline}
\small
\setlength{\tabcolsep}{5pt}
\renewcommand{\arraystretch}{1.12}
\begin{tabularx}{0.88\linewidth}{l*{2}{>{\centering\arraybackslash}X}}
    \toprule
    Component & PyTorch CUDA (s)$\downarrow$ & Compiled CUDA (s)$\downarrow$ \\
    \midrule
    Prism warp                    & 0.360 & 0.356 \\
    Threshold-event generation   & 0.753 & 0.125 \\
    Timestamp-aware compensation & 0.047 & 0.045 \\
    \midrule
    \rowcolor{ourgray}
    Total                         & 1.162 & \textbf{0.528} \\
    \bottomrule
\end{tabularx}
\end{table}

\end{document}